\documentclass[journal]{IEEEtran}

\ifCLASSINFOpdf

\else

\fi

\usepackage{orcidlink}
\usepackage{caption}
\usepackage{cite}
\usepackage{cuted}
\usepackage{array}
\usepackage{amsmath,amssymb,amsfonts}
\usepackage{amsthm}
\usepackage{makecell}
\usepackage{algorithmic}
\usepackage{graphicx}
\usepackage{subcaption}
\usepackage{amsmath}
\usepackage{textcomp}
\usepackage{xcolor}
\usepackage{float}
\usepackage{stfloats}
\usepackage{multirow}
\usepackage{threeparttable}
\newtheorem{theorem}{\bf {Theorem}}
\def\BibTeX{{\rm B\kern-.05em{\sc i\kern-.025em b}\kern-.08em
    T\kern-.1667em\lower.7ex\hbox{E}\kern-.125emX}}

\begin{document}

\title{{KAN versus MLP on Irregular or Noisy Functions}\\

\author{
        Chen~Zeng,
        Jiahui~Wang,
        Haoran~Shen,
    and~Qiao~Wang\orcidlink{0000-0002-5271-0472},~\IEEEmembership{Senior Member,~IEEE}%
 
\thanks{Both  C. Zeng and H. Shen was with the School of Information Science and Engineering, Southeast University, Nanjing, China (email: haoranshen28@163.com, zengchen0797@163.com). }
\thanks{J. Wang was with the School of Economics and Management, Southeast University, Nanjing, China (email: wangjh0512@qq.com).}
\thanks {Q. Wang was with both the School of Information Science and Engineering and the School of Economics and Management, Southeast University, Nanjing, China (Corresponding Author, email: qiaowang@seu.edu.cn).}  
}}

\maketitle

\begin{abstract}
In this paper, we compare the performance of Kolmogorov-Arnold Networks (KAN) and Multi-Layer Perceptron (MLP) networks on irregular or noisy functions. We control the number of parameters and the size of the training samples to ensure a fair comparison. For clarity, we categorize the functions into six types: regular functions, continuous functions with local non-differentiable points, functions with jump discontinuities, functions with singularities, functions with coherent oscillations, and noisy functions. Our experimental results indicate that KAN does not always perform best. For some types of functions, MLP outperforms or performs comparably to KAN. Furthermore, increasing the size of training samples can improve performance to some extent. When noise is added to functions, the irregular features are often obscured by the noise, making it challenging for both MLP and KAN to extract these features effectively. We hope these experiments provide valuable insights for future neural network research and encourage further investigations to overcome these challenges.
\end{abstract}


\section{Introduction}
The Kolmogorov-Arnold networks (KAN) have garnered significant attention since their appearance on arXiv \cite{liu2024kan}. These networks utilize the Kolmogorov-Arnold representation theorem, which posits that any multivariate continuous function can be expressed as a combination of continuous single-variable functions and addition. Unlike conventional Multi-Layer Perceptron (MLP) networks, KANs incorporate learnable activation functions. According to \cite{liu2024kan}, this feature provides KANs with enhanced interpretability and accuracy over MLPs.

Numerous investigations into KAN applications have rapidly surfaced, covering areas such as smart energy grid optimization\cite{wang2024blackboxclarityaipowered}\cite{2024arXiv240800273T}, chemistry data analysis\cite{2024arXiv240616026W}\cite{2024arXiv240720265L}, image classification\cite{2024arXiv240600600C}\cite{2024arXiv240607869T}\cite{2024arXiv240716268I}, deep function learning\cite{2024arXiv240704819Z}, quantum architecture search\cite{2024arXiv240617630K}, medical image analysis and processing\cite{2024arXiv240602918L}\cite{2024arXiv240609931C}, disease risk predictions\cite{2024arXiv240706560D}, graph learning tasks\cite{2024arXiv240606470K}\cite{2024arXiv240801018L}\cite{2024arXiv240713044G}, asset pricing models\cite{2024arXiv240802694W}, 3D object detection (in autonomous driving)\cite{2024arXiv240802088L}, sentiment analysis\cite{2024arXiv240710347L}, and deep kernel learning\cite{2024arXiv240721176Z}.

On the contrary, a growing body of research has highlighted the imperfections of KANs compared to MLPs. For example, \cite{Zhang2024} and our work \cite{our_2024kan_With_noise} noted the vulnerability of KANs to noise, indicating that even minor noise can lead to a significant rise in test loss. Additionally, \cite{limitationclassification} claimed that KANs do not outperform MLPs in highly complex datasets and require considerably more hardware resources. Furthermore, \cite{yu2024kanmlpfairercomparison} noted that MLPs generally have higher accuracy than KANs across various standard machine learning tasks, with the exception of tasks involving symbolic formula representation.

Moreover, it is widely recognized that the regularity of the functions being approximated significantly influences the efficacy of neural networks. Functions that are smooth and continuous, known as regular functions, are generally approximated more precisely by KANs. On the other hand, functions that exhibit discontinuities or abrupt variations, referred to as irregular functions, present a greater difficulty.

 It is essential to emphasize that the research by \cite{yu2024kanmlpfairercomparison} focused on achieving a fair comparison between KANs and MLPs, yet it excluded the effect of irregularities or noise.  In this investigation, we assess the performance of MLP and KAN in modeling irregular or noisy functions. To ensure fairness, we control the number of parameters and the amount of training data. Moreover, we investigate the influence of different optimizers on the accuracy of fitting specific functions. This research continues directly and naturally from our recent study on the efficacy of KANs in fitting noisy functions \cite{our_2024kan_With_noise}.

The structure of this paper is organized as follows: Section \ref{KAN} provides an introduction to the Kolmogorov-Arnold Theorem and KANs, discussing their benefits and limitations, and enumerates the six types of functions. Section \ref{Cmp_ir} evaluates the performance of MLP and KAN in approximating regular and irregular functions. KAN performs better than MLP in the case of regular functions, whereas for certain irregular functions, MLP shows superior performance over KAN. Increasing the number of training samples can enhance performance to a certain degree. Moreover, the impact of different optimizers on specific functions is examined. Adam generally performs better than L-BFGS during extensive training on most functions. Section \ref{Cmp_noise} introduces noise to the previously utilized functions and continues the comparison between MLP and KAN. It is observed that noise can obscure the characteristics of irregular components of functions, making it difficult for both MLP and KAN to capture these features accurately. Finally, Section \ref{Conclusion} summarizes the findings of our experiments.

\section{Komogorov-Arnold Theorem and KANs}\label{KAN}

\begin{table*}[htbp]
\label{tab-1}
\renewcommand\arraystretch{1.7}
\caption{Several Types of Functions and Their Examples}
\centering
\begin{tabular}{|c|cc|}
\hline
\multirow{3}{*}{\makecell{\\\\\\\\\\Regular}}    & \multicolumn{2}{c|}{Smooth}                                                          \\ \cline{2-3} 
                            & \multicolumn{1}{c|}{$\qquad\qquad\   f_1(x)=x^2 
                            \qquad\qquad\    $}                               & {$f_2(x)=e^x$}                             \\
                            & \multicolumn{1}{c|}{\includegraphics[width=0.15\linewidth]{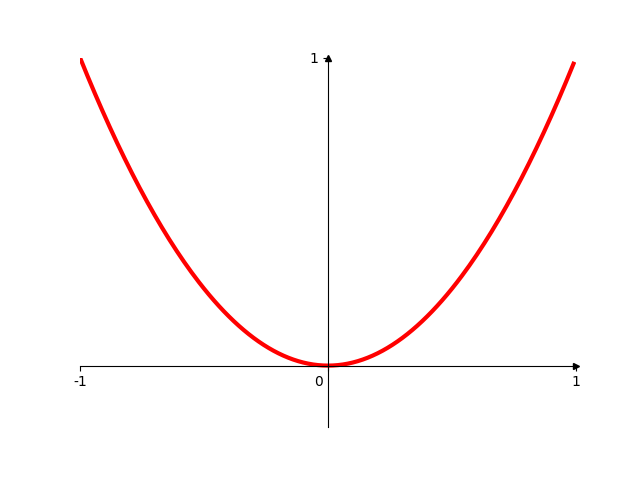}}                               & {\includegraphics[width=0.15\linewidth]{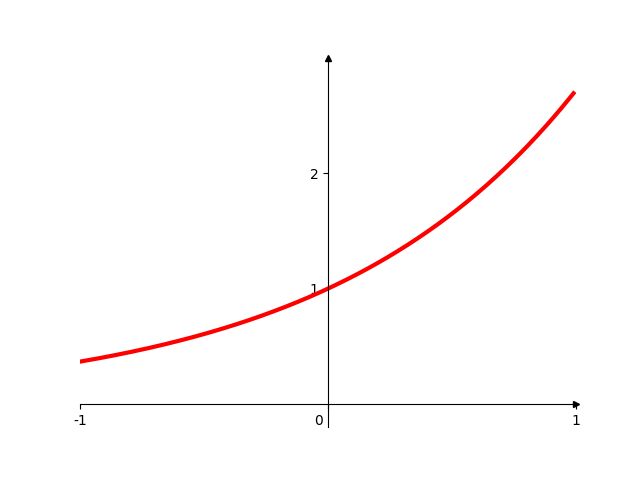}}                             \\ \hline
\multirow{12}{*}{\makecell
{\\\\\\\\\\\\\\\\\\\\\\\\\\\\\\\\\\\\\\\\\\Irregular}} & \multicolumn{2}{c|}{Continuous everywhere except at points of non-differentiability} \\ \cline{2-3} 
                            & \multicolumn{1}{c|}{$f_3(x)=|x|$}                               & {$f_4(x)=1-\sqrt{|x|}$}                             \\
                            & \multicolumn{1}{c|}{\includegraphics[width=0.15\linewidth]{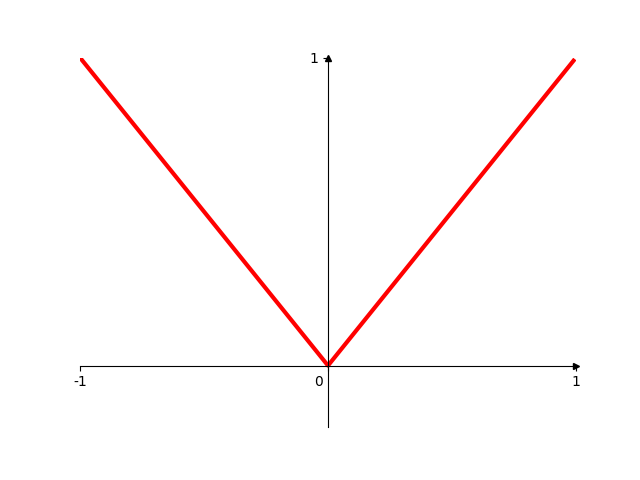}}                               & {\includegraphics[width=0.15\linewidth]{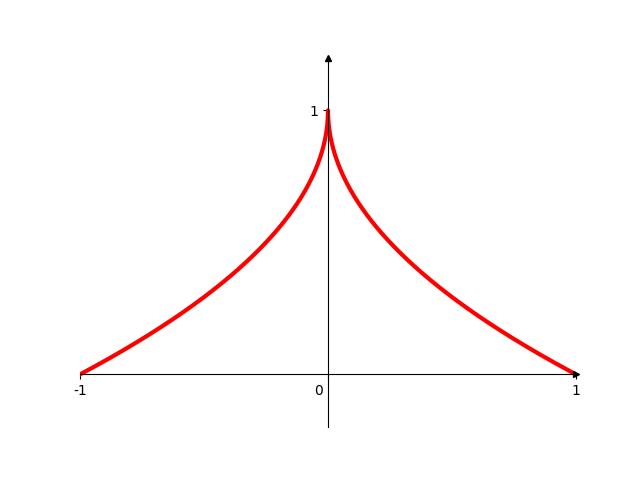}}                             \\ \cline{2-3} 
                            & 
\multicolumn{2}{c|}{Jump}                                                            \\ \cline{2-3} 
                            & \multicolumn{1}{c|}{$f_5(x)=\begin{cases}
1,\;|x|<0.5\\
0,\;\text{other}\\
\end{cases}$}                               & {$f_6(x)=\begin{cases}
1-4x^2,\;|x|<0.5\\
1,\;\text{other}\\
\end{cases}$}                             \\
                            & \multicolumn{1}{c|}{\includegraphics[width=0.15\linewidth]{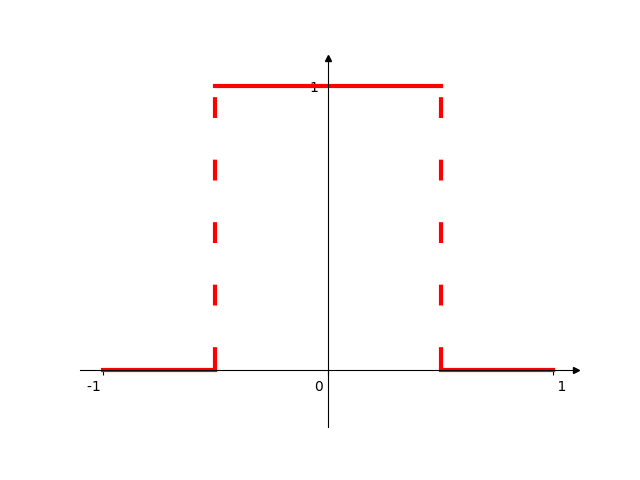}}                               & {\includegraphics[width=0.15\linewidth]{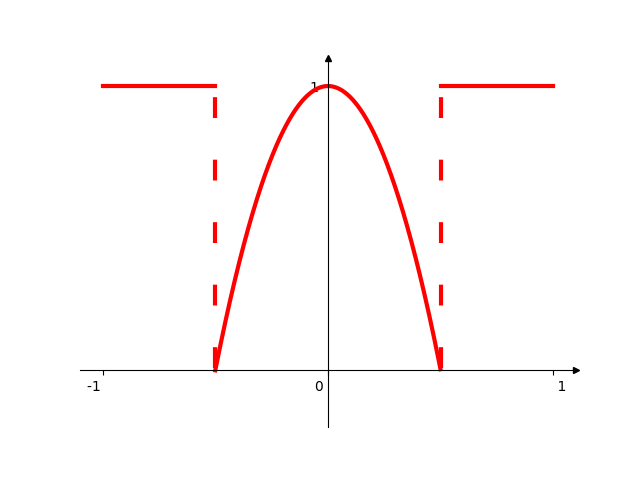}}                             \\ \cline{2-3} 
                            & 
\multicolumn{2}{c|}{Singular}                                                        \\ \cline{2-3} 
                            & \multicolumn{1}{c|}{$f_7(x)=\frac{1}{x}$}                               & {$f_8(x)=\frac{1}{1-x^2}-1$}                             \\
                            & \multicolumn{1}{c|}{\includegraphics[width=0.15\linewidth]{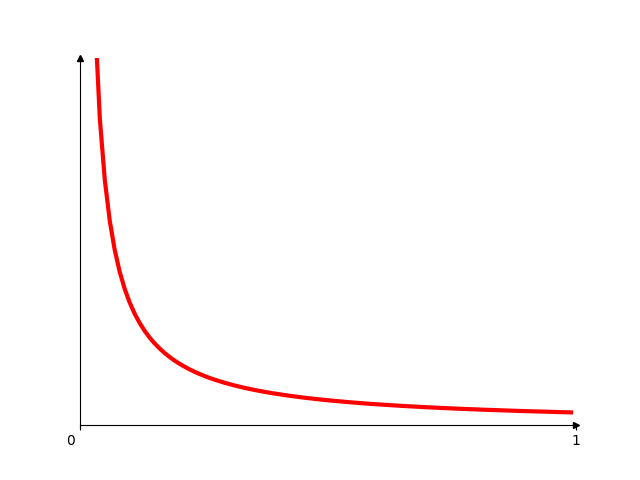}}                               & {\includegraphics[width=0.15\linewidth]{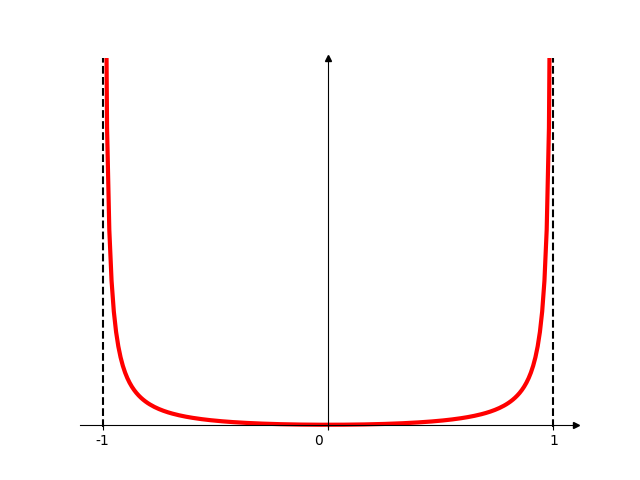}}                             \\ \cline{2-3} 
                            & 
\multicolumn{2}{c|}{Coherent oscillation}                                            \\ \cline{2-3} 
                            & \multicolumn{1}{c|}{$f_9(x)=\cos (\frac{1}{x})$}                               & {$f_{10}(x)=\cos (\frac{2\pi}{1-x^2})$}                             \\
                            & \multicolumn{1}{c|}{\includegraphics[width=0.15\linewidth]{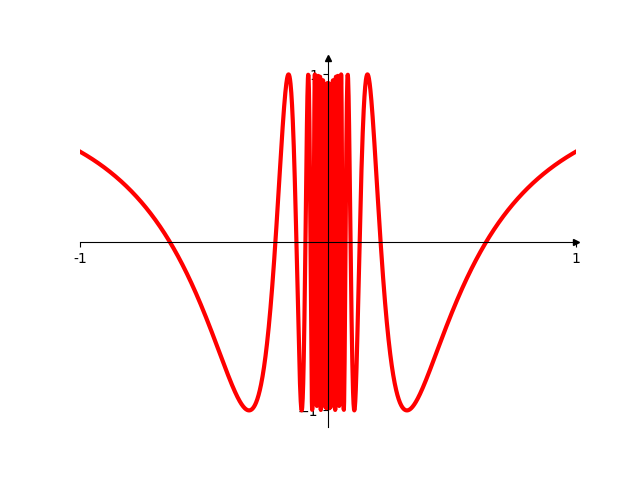}}                               & {\includegraphics[width=0.15\linewidth]{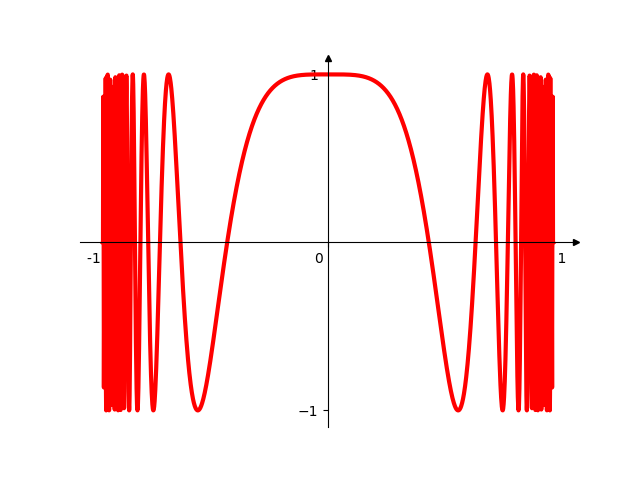}}                             \\ \hline
\multirow{3}{*}{\makecell{\\\\\\\\\\\\Noisy}}      & 
\multicolumn{2}{c|}{Noisy}                                                           \\ \cline{2-3} 
                            & \multicolumn{2}{c|}{$y=x+n(x)$, where $n(x)$ denotes additive noise.}                                                             \\
                            & \multicolumn{2}{c|}{\includegraphics[width=0.3\linewidth]{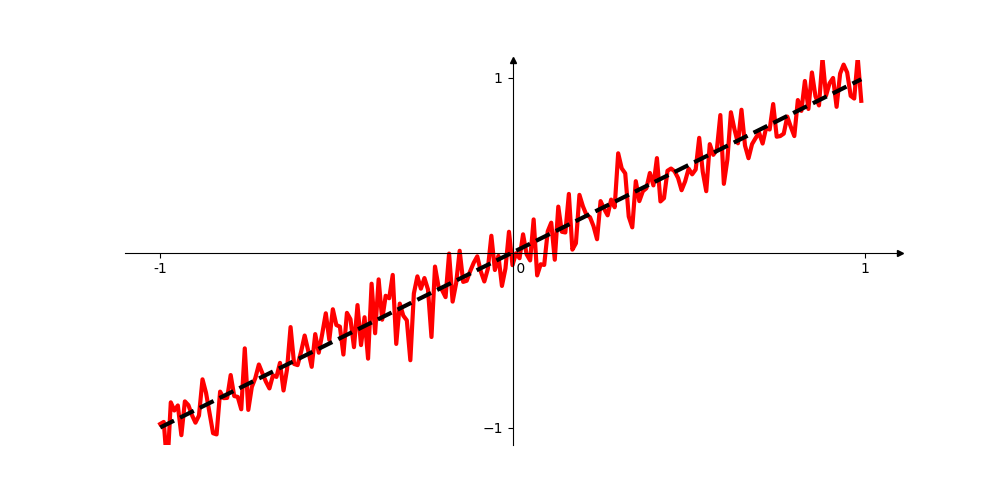}}                                                             \\ \hline
\end{tabular}
\label{tab-1}
\end{table*}

The Kolmogorov-Arnold theorem pertains to expressing multivariable continuous functions. According to the theorem, any continuous function involving multiple variables can be expressed as a combination of continuous single-variable functions and addition \cite{Kolmogorov1956-2} \cite{Kolmogorov1957} \cite{Arnold1957}. Formally, it can be stated as:

\begin{theorem}\label{th:1}
[Kolmogorov-Arnold Theorem]
Let $f:\ [0,1]^n\to \mathbb R$ be any multivariate continuous function, there exist continuous univariate functions \( \phi_i \) and \( \psi_{ij} \) such that:
\begin{equation}
f(x_1, x_2, \ldots, x_n) = \sum_{i=1}^{2n+1} \phi_i \left( \sum_{j=1}^{n} \psi_{ij}(x_j) \right).\label{ka1}
\end{equation}\label{ka-1}
\end{theorem}
Leveraging the Kolmogorov-Arnold theorem, KANs introduce a novel neural network architecture. Unlike traditional Multi-Layer Perceptrons (MLPs) which use fixed activation functions, KANs employ flexible activation functions. Each weight in the network is replaced by a univariate function, often modeled as a spline function. This methodology is theoretically advantageous in enhancing the adaptability of KANs across different datasets and applications.

Unfortunately, Theorem \ref{th:1} was proven using a non-constructive approach, without providing a constructive proof. In 2009, \cite{Braun2009} presented a constructive proof for this theorem. However, issues can arise when dealing with functions that show irregular behaviors. In mathematical analysis, it is common to classify at least five distinct types of irregularities. Details and examples of these types are presented in Table \ref{tab-1}.

\section{Comparison on irregular functions}\label{Cmp_ir}

Here, we offer some explanations for the initial five types mentioned in Table \ref{tab-1}. We compare these functions using multiple sets of KAN and MLP networks that have similar parameter counts. The parameter numbers for each network are presented in Table \ref{network}. We implement the L-BFGS optimizer for these functions as it shows better performance in small-scale training. For functions with singularities or coherent oscillations, which might need more training samples and iterations, we also investigate the Adam optimizer's capability.

\begin{table}[htbp]
\renewcommand\arraystretch{1.2}
\caption{The number of parameters for each KAN and MLP network.}
\centering
\begin{tabular}{|cc|cccc|}
\hline
\multicolumn{2}{|c|}{MLP}                      & \multicolumn{4}{c|}{KAN}                                \\ \hline
\multicolumn{1}{|c|}{width}        & parameter & width       & grid & \multicolumn{1}{c|}{k} & parameter \\ \hline
\multicolumn{1}{|c|}{{[}1,7,1{]}}  &   22        & {[}1,1,1{]} & 3    & \multicolumn{1}{c|}{3} &   24        \\ \hline
\multicolumn{1}{|c|}{{[}1,39,1{]}} &   118        & {[}1,5,1{]} & 3    & \multicolumn{1}{c|}{3} &   120        \\ \hline
\multicolumn{1}{|c|}{{[}1,79,1{]}} &   238        & {[}1,10,1{]} & 3    & \multicolumn{1}{c|}{3} &    240       \\ \hline
\end{tabular}
\label{network}
\end{table}

\subsection{Regular functions} 

\begin{figure*}[htbp]
  \centering
  \subfloat[\centering
  $f_1$
  \label{f1_num}]
  {
    \includegraphics[width=0.45\linewidth]{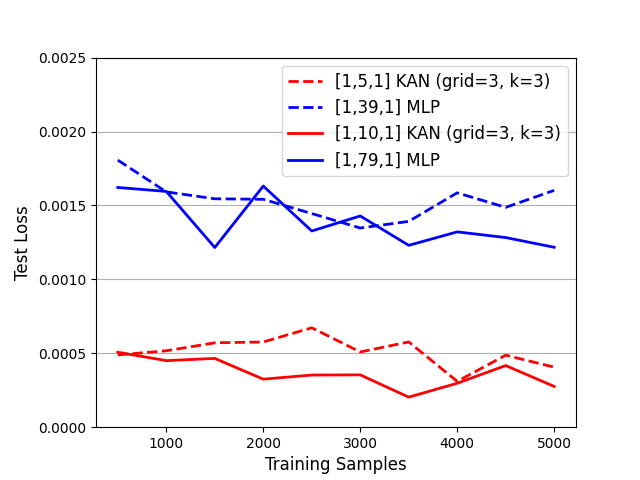}
  }
  \subfloat[\centering
  $f_2$
  \label{f2_num}]
  {
    \includegraphics[width=0.45\linewidth]{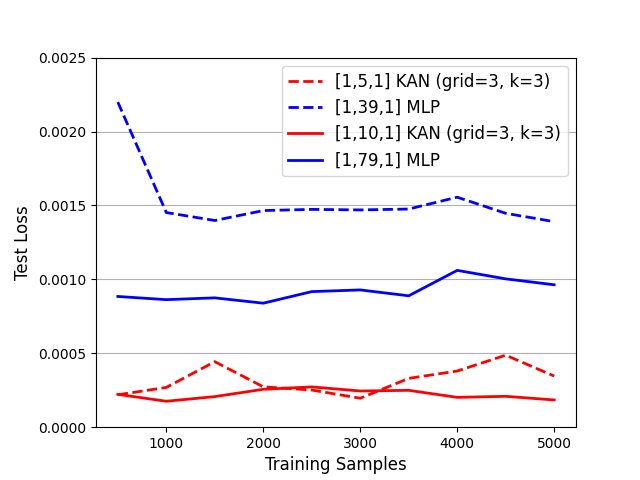}
  }
  \caption{Recover $f_1$ and $f_2$ independently using KAN and MLP  under different training sample sizes.}
  \label{type0}
\end{figure*}

First, consider the functions exhibiting strong regularity. Such functions are continuous and differentiable at all points, exhibiting no discontinuities or non-differentiable regions, similar to $f_1$ and $f_2$. We reconstruct these two functions using two sets of MLP and KAN networks that have comparable parameters but are trained with different sample sizes. The outcomes are displayed in Fig \ref{type0}. It can be observed that for this category of functions, KAN outperforms MLP.

\subsection{Continuous functions with points where derivatives do not exist}

The functions $f_3$ and $f_4$ serve as prime examples of this category. They maintain continuity at all points except at $x = 0$, where they are non-differentiable. Likewise, these functions are retrieved individually using independent MLP and KAN network sets.

\begin{figure*}[htbp]
  \centering
  \subfloat[\centering
  Various epochs, $f_3$,  
  training samples = 50
  \label{f3_epoch}]
  {
    \includegraphics[width=0.23\linewidth]{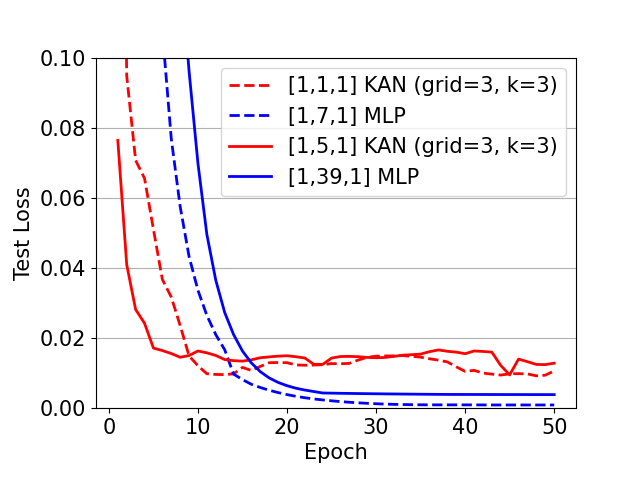}
  }
  \subfloat[\centering
  Various epochs, $f_4$, 
  training samples = 50
  \label{f4_epoch}]
  {
    \includegraphics[width=0.23\linewidth]{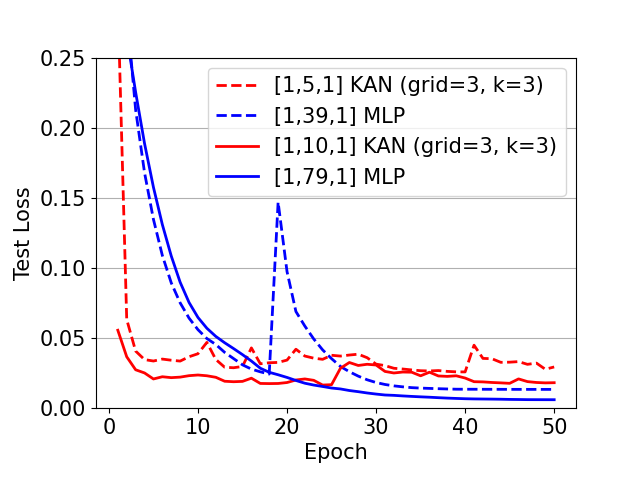}
  }
  \subfloat[\centering
  Various training samples, $f_3$
  \label{f3_num}]
  {
    \includegraphics[width=0.23\linewidth]{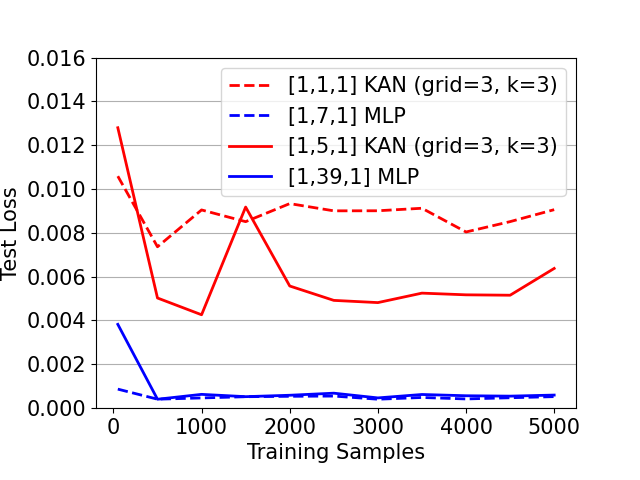}
  }
  \subfloat[\centering
  Various training samples, $f_4$
  \label{f4_num}]
  {
    \includegraphics[width=0.23\linewidth]{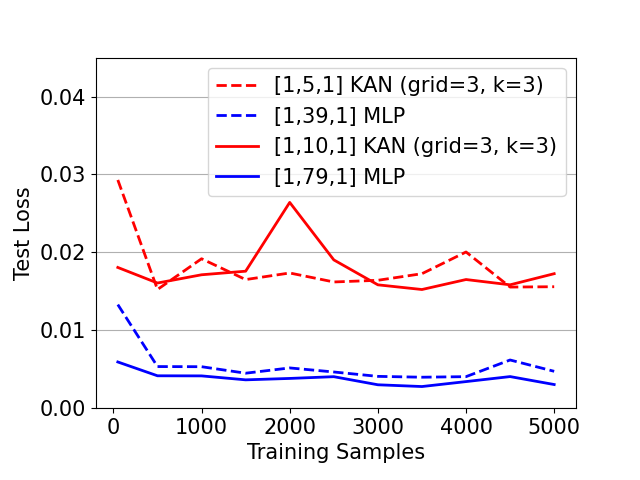}
  }

  \subfloat[\centering
  Real values and 
  predictions of MLP, 
  $f_3$, training samples = 50
  \label{f3_mlp_50}]
  {
    \includegraphics[width=0.23\linewidth]{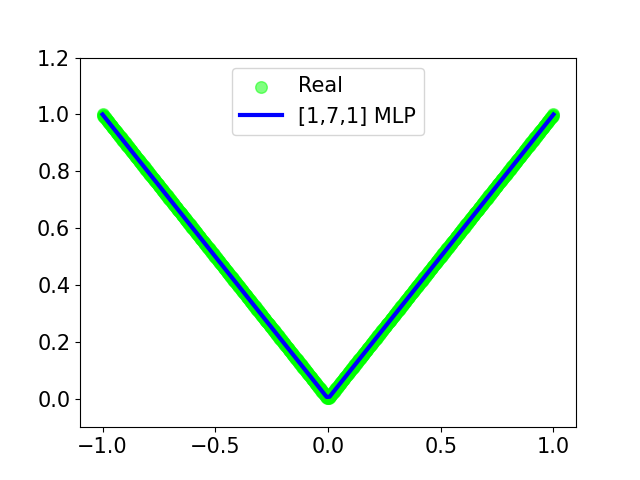}
  }
  \subfloat[\centering
  Real values and 
  predictions of KAN, 
  $f_3$, training samples = 50
  \label{f3_kan_50}]
  {
    \includegraphics[width=0.23\linewidth]{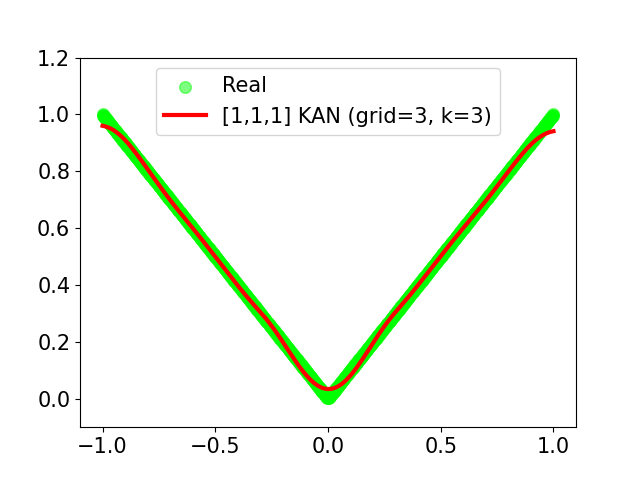}
  }
  \subfloat[\centering
  Real values and 
  predictions of MLP, 
  $f_3$, training samples = 5000
  \label{f3_mlp_5000}]
  {
    \includegraphics[width=0.23\linewidth]{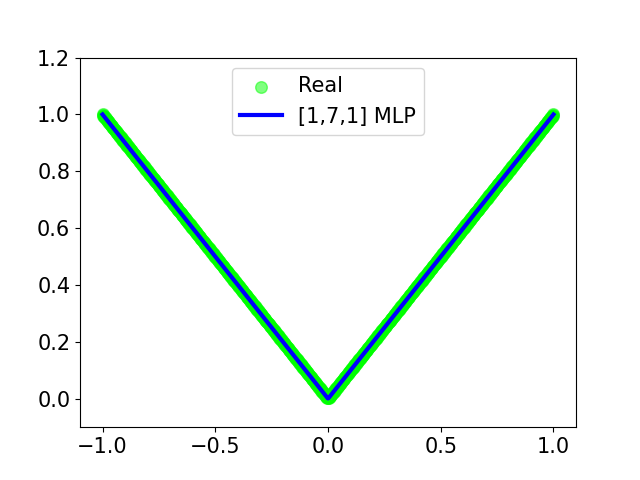}
  }
  \subfloat[\centering
  Real values and 
  predictions of KAN, 
  $f_3$, training samples = 5000
  \label{f3_kan_5000}]
  {
    \includegraphics[width=0.23\linewidth]{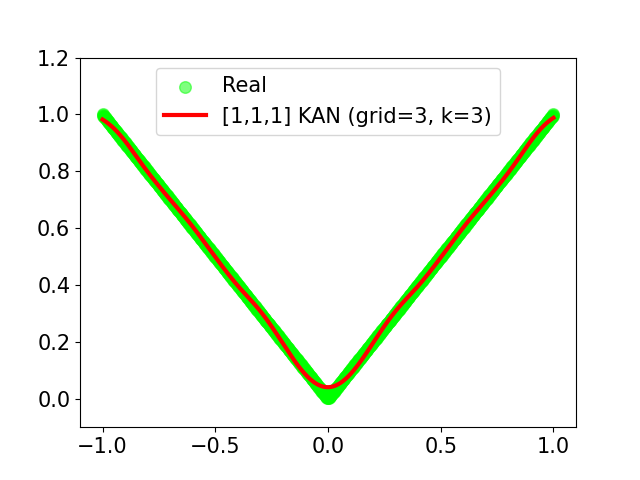}
  }

  \subfloat[\centering
  Real values and 
  predictions of MLP, 
  $f_4$, training samples = 50
  \label{f4_mlp_50}]
  {
    \includegraphics[width=0.23\linewidth]{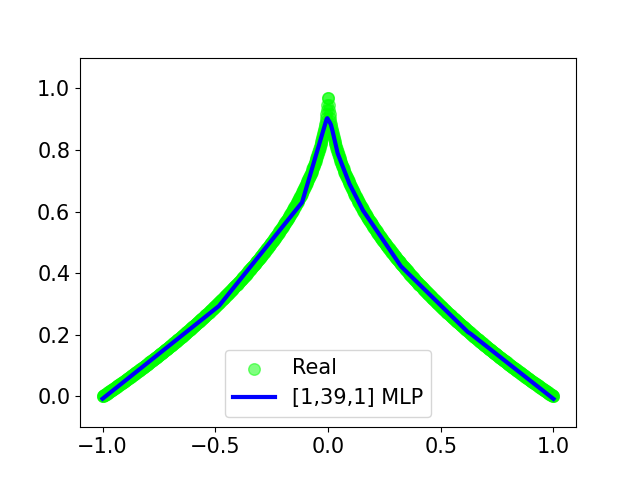}
  }
  \subfloat[\centering
  Real values and 
  predictions of KAN, 
  $f_4$, training samples = 50
  \label{f4_kan_50}]
  {
    \includegraphics[width=0.23\linewidth]{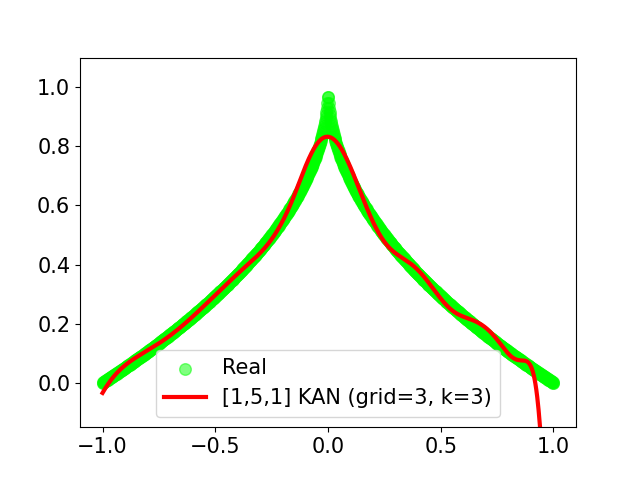}
  }
  \subfloat[\centering
  Real values and 
  predictions of MLP, 
  $f_4$, training samples = 5000
  \label{f4_mlp_5000}]
  {
    \includegraphics[width=0.23\linewidth]{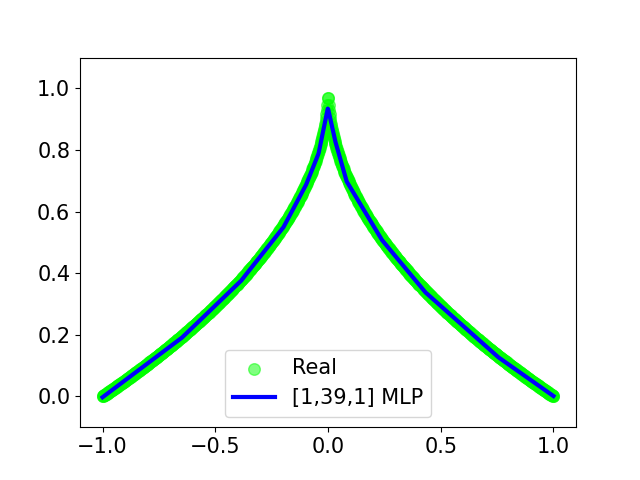}
  }
  \subfloat[\centering
  Real values and 
  predictions of KAN, 
  $f_4$, training samples = 5000
  \label{f4_kan_5000}]
  {
    \includegraphics[width=0.23\linewidth]{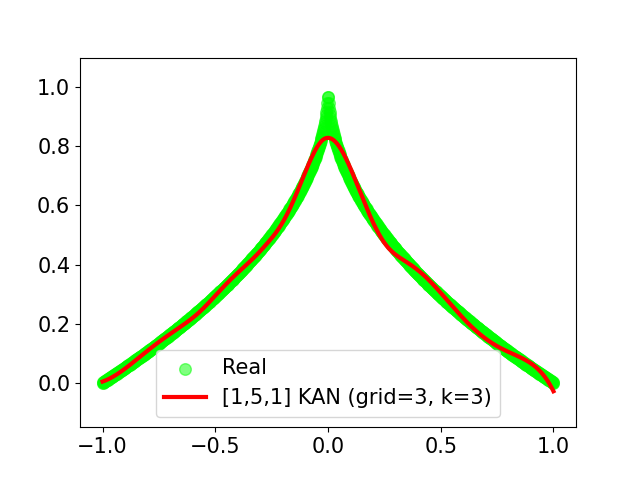}
  }
  \caption{Recover $f_3$ and $f_4$ independently using KAN and MLP.}
  \label{type1}
\end{figure*}

The outcomes are illustrated in Fig \ref{type1}. For these particular functions, the KAN's performance is worse than the MLP's. Despite the MLP network's slower convergence, it eventually reaches a lower test loss. Additionally, it can be noted that amplifying the training sample size marginally enhances the performance of both networks. However, in the vicinity of the non-differentiable point, the MLP shows more significant improvement than the KAN. More visually, it is evident that the fitting performance of MLP and KAN around $x = 0$ (the non-differentiable point) is approximately the same. Yet, with a larger training sample size, the MLP demonstrates superior fitting performance near $x = 0$ compared to the KAN.

\subsection{Functions with jump}

\begin{figure*}[htbp]
  \centering
  \subfloat[\centering
  Various epochs, $f_5$, 
  training samples = 50
  \label{f5_epoch}]
  {
    \includegraphics[width=0.23\linewidth]{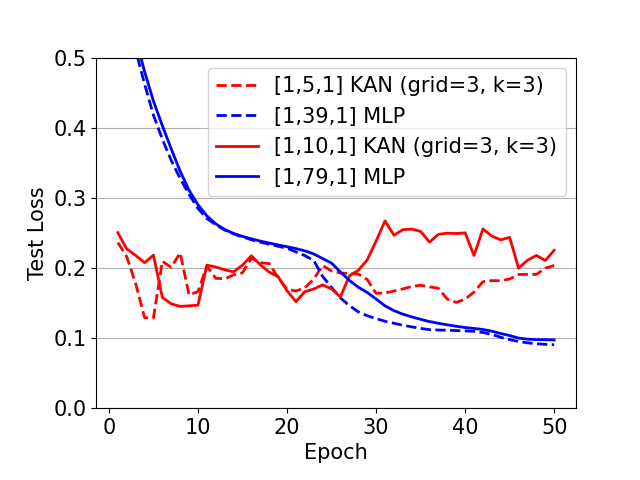}
  }
  \subfloat[\centering
  Various training samples, $f_5$
  \label{f5_num}]
  {
    \includegraphics[width=0.23\linewidth]{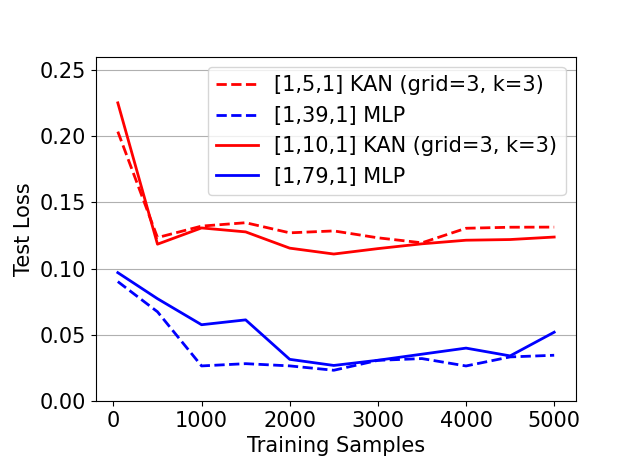}
  }
  \subfloat[\centering
  Real values and predictions, $f_5$, training samples = 50
  \label{f5_50}]
  {
    \includegraphics[width=0.23\linewidth]{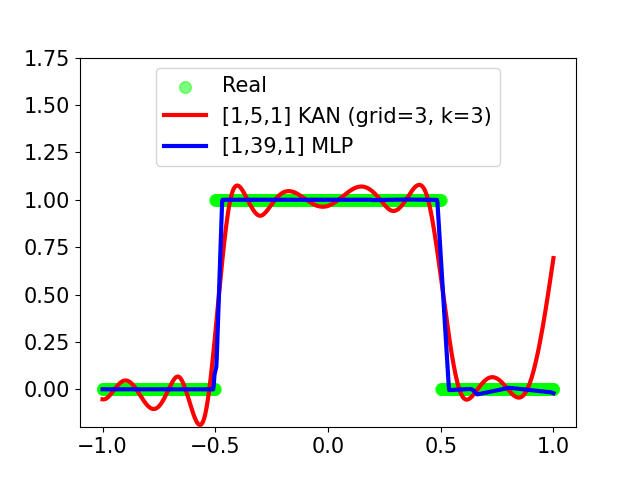}
  }
  \subfloat[\centering
  Real values and predictions, $f_5$, training samples = 5000
  \label{f5_5000}]
  {
    \includegraphics[width=0.23\linewidth]{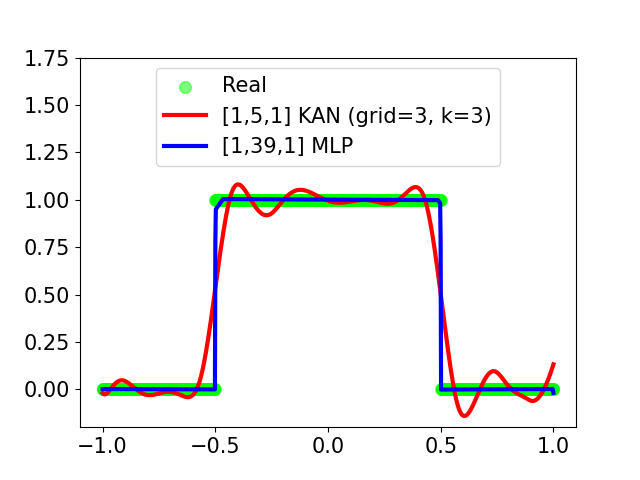}
  }

  \subfloat[\centering
  Various epochs, $f_6$, 
  training samples = 50
  \label{f6_epoch}]
  {
    \includegraphics[width=0.23\linewidth]{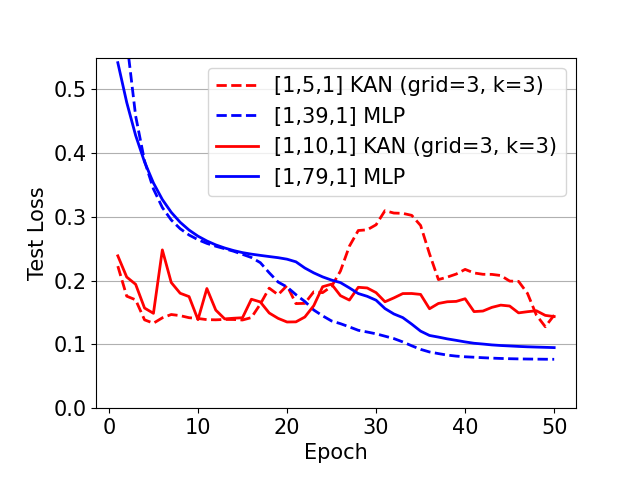}
  }
  \subfloat[\centering
  Various training samples, $f_6$
  \label{f6_num}]
  {
    \includegraphics[width=0.23\linewidth]{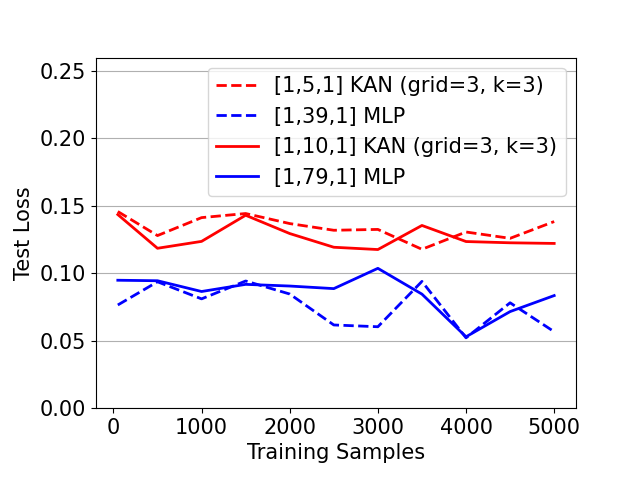}
  }
  \subfloat[\centering
  Real values and predictions, $f_6$, training samples = 50
  \label{f6_50}]
  {
    \includegraphics[width=0.23\linewidth]{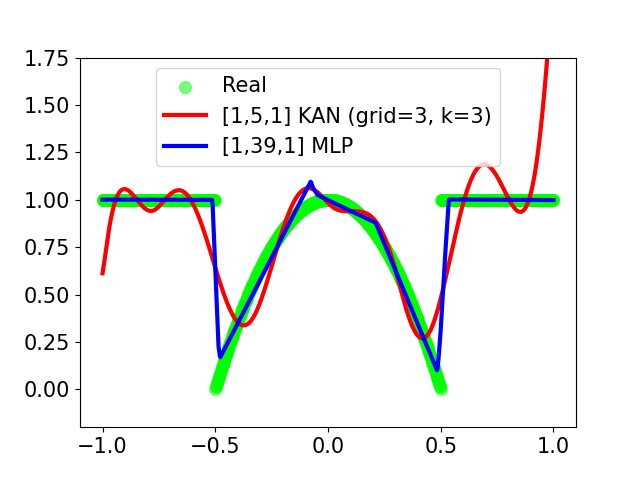}
  }
  \subfloat[\centering
  Real values and predictions, $f_6$, training samples = 5000
  \label{f6_5000}]
  {
    \includegraphics[width=0.23\linewidth]{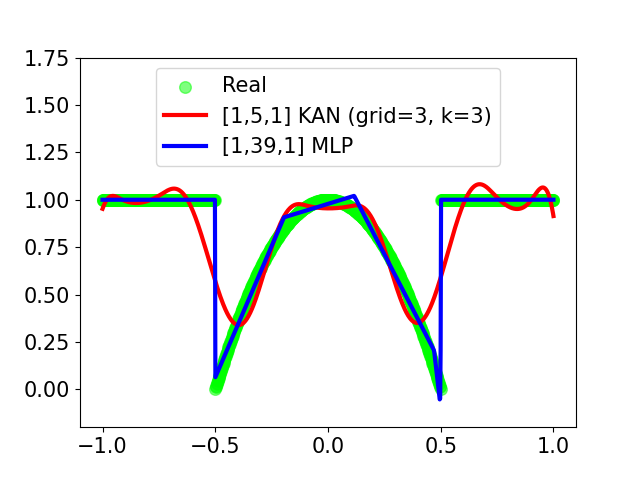}
  }
  \caption{Recover $f_5$ and $f_6$ independently using KAN and MLP.}
  \label{type2}
\end{figure*}

The examples of this category include $f_5$ and $f_6$. These functions have jump discontinuities at $x = \pm 0.5$, where the function values abruptly change between $0$ and $1$. The experimental outcomes for these functions are depicted in Fig \ref{type2}. Similar to the earlier type, the results show that the MLP outperforms the KAN. Moreover, expanding the training dataset size can enhance both networks' performance to a certain extent. Nevertheless, KAN consistently fails to match the performance of MLP.

\subsection{Functions with singularities}

Functions possessing singularities display distinct behaviors, marked by a rapid change rate as they near these points, with the absolute value of their first derivative tending towards positive infinity at the singularity. Additionally, for any chosen continuous interval that omits these singularities, the functions remain continuous and differentiable across the interval.

To avoid division by zero and guarantee clear fitting results, the ranges of the functions $f_7$ and $f_8$ are limited to $[0.001, 1]$ and $[-0.999, 0.999]$, respectively.

\begin{figure*}[htbp]
  \centering
  \subfloat[\centering
  $f_7$
  \label{f7_trainingsamples}]
  {
    \includegraphics[width=0.47\linewidth]{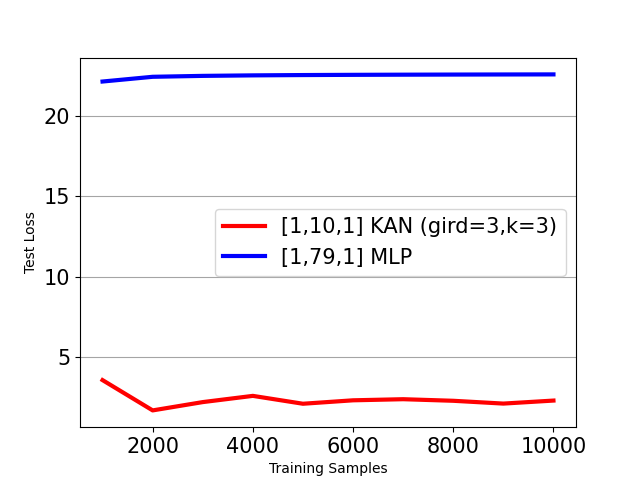}
  }
  \subfloat[\centering
  $f_8$
  \label{f8_trainingsamples}]
  {
    \includegraphics[width=0.47\linewidth]{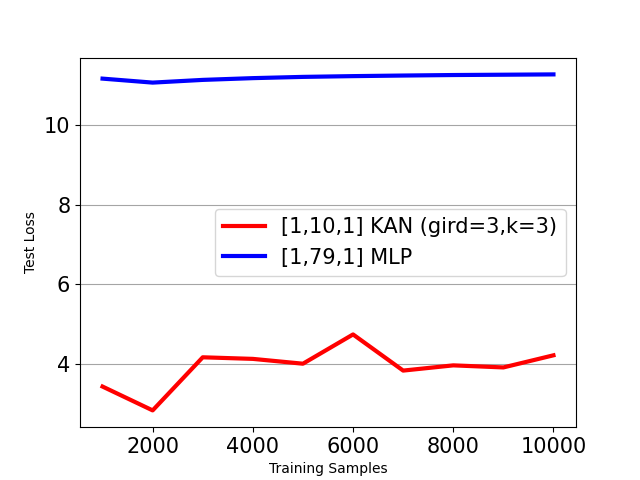}
  }

  \caption{Recover $f_7$ and $f_8$ independently using KAN and MLP with optimizer Adam, 2000 Epochs}
  \label{type3_trainingsamples}
\end{figure*}

Before performing specific fitting experiments, we examined the effects of the training sample size, the number of Epochs, and the selection of optimizer on the fitting performance. As illustrated in Fig \ref{type3_trainingsamples}, simply enlarging the sample size by itself does not substantially enhance the performance when recovering $f_7$ and $f_8$.

\begin{figure*}[htbp]
  \centering
  \subfloat[\centering
  MLP, $f_7$
  \label{f7_MLP_Opt}]
  {
    \includegraphics[width=0.23\linewidth]{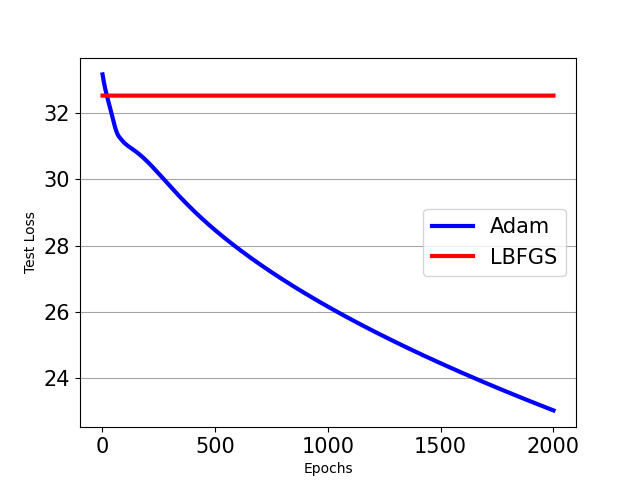}
  }
  \subfloat[\centering
  KAN, $f_7$
  \label{f7_KAN_Opt}]
  {
    \includegraphics[width=0.23\linewidth]{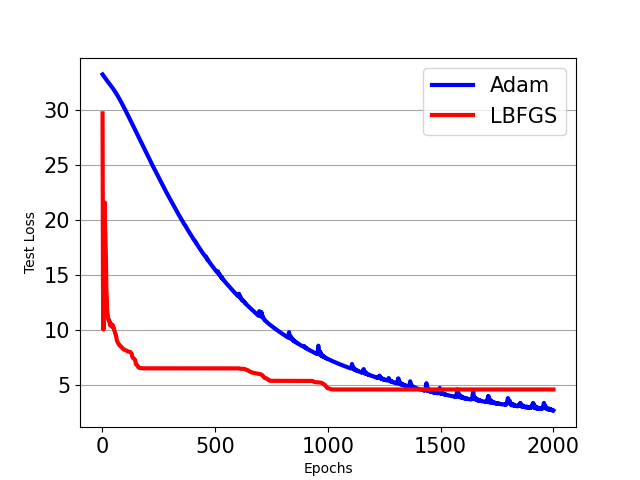}
  }
  \subfloat[\centering
  MLP, $f_8$
  \label{f8_MLP_Opt}]
  {
    \includegraphics[width=0.23\linewidth]{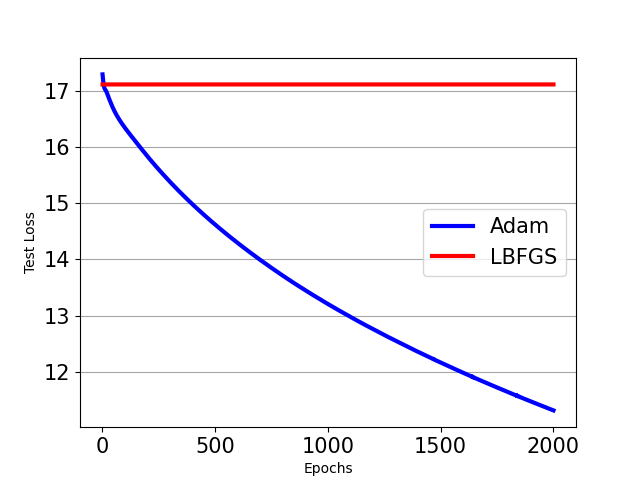}
  }
  \subfloat[\centering
  KAN, $f_8$
  \label{f8_KAN_Opt}]
  {
    \includegraphics[width=0.23\linewidth]{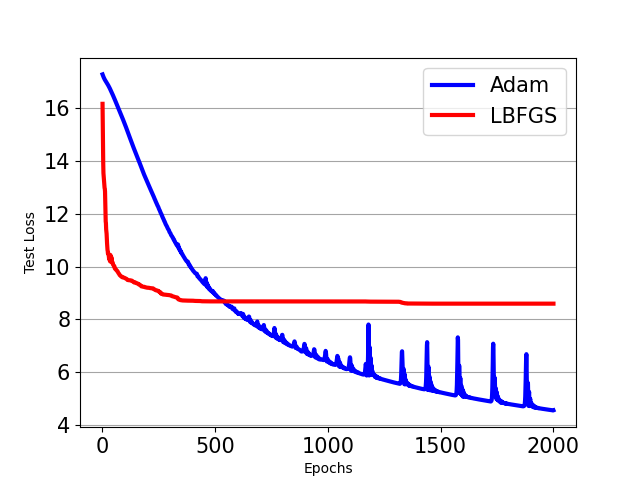}
  }

  \caption{The variation of test loss with the increasing number of epochs when recovering $f_7$ and $f_8$ independently using [1,10,1] KAN (grid=3,k=3) and [1,79,1] MLP with optimizer L-BFGS and Adam, learning rate=0.01}
  \label{type3_Opt}
\end{figure*}

Pykan offers two optional optimizers: Adam and L-BFGS. As depicted in Fig \ref{type3_Opt}, while the L-BFGS optimizer achieves faster convergence with fewer epochs, the Adam optimizer shows superior performance in test loss with an increased number of epochs, particularly within MLP networks. This suggests that using the Adam optimizer and increasing the number of epochs can be an effective method to improve fitting performance. However, it is crucial to recognize that with a fixed learning rate, the improvement from this strategy is naturally constrained.

An additional limitation of the L-BFGS algorithm is its substantial time complexity. As illustrated in Table \ref{tab_t3_Opt}, for an identical number of epochs, the training duration of the network with the L-BFGS optimizer is frequently several times greater than with the Adam optimizer.

\begin{figure*}[htbp]
  \centering
  \subfloat[\centering
  $f_7$, MLP
  \label{f7_mlp}]
  {
    \includegraphics[width=0.45\linewidth]{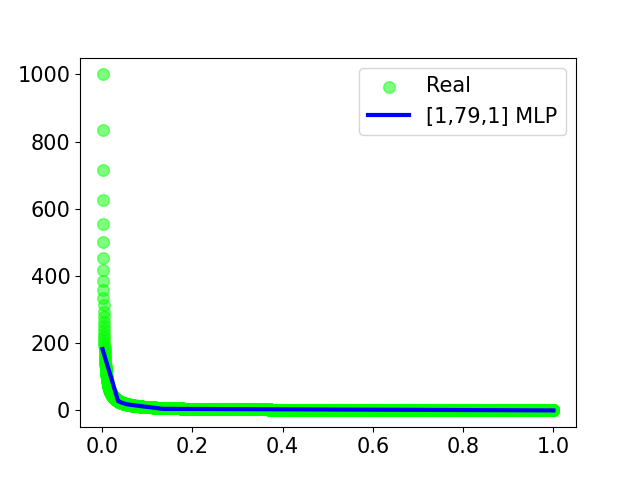}
  }
  \subfloat[\centering
  $f_7$, KAN 
  \label{f7_kan}]
  {
    \includegraphics[width=0.45\linewidth]{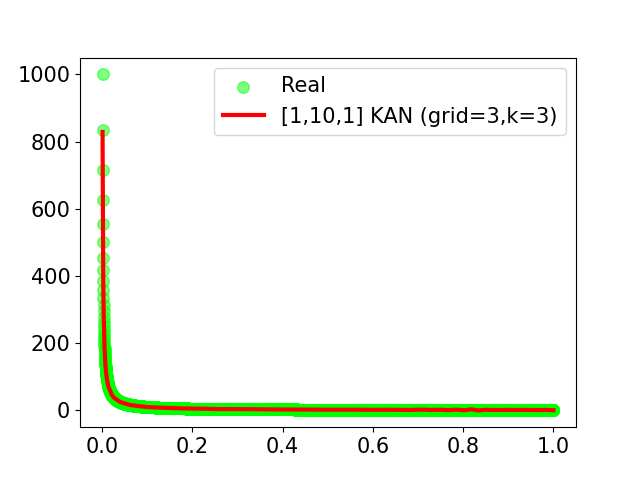}
  }
  
  \subfloat[\centering
  $f_8$, MLP
  \label{f8_mlp}]
  {
    \includegraphics[width=0.45\linewidth]{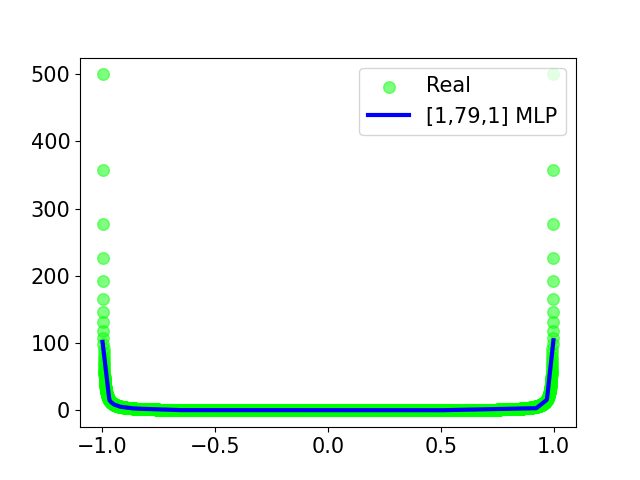}
  }
  \subfloat[\centering
  $f_8$, KAN
  \label{f8_kan}]
  {
    \includegraphics[width=0.45\linewidth]{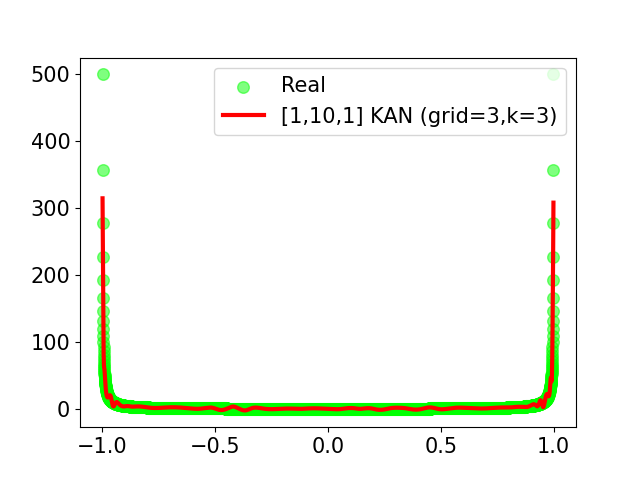}
  }

  \caption{Recover $f_7,(x\in [0.001,1])$ and $f_8,(x\in[-0.999,0.999])$ independently using [1,10,1] KAN (grid=3, k=3) and [1,79,1] MLP both with optimizer Adam, same or different epochs}
  \label{type3_fit}
\end{figure*}

Drawing from earlier experiments and findings, the fitting tests will utilize the Adam optimizer set with a learning rate of 0.01. As shown in Fig. \ref{type3_fit}, KAN outperformed MLP in terms of fitting functions with singularities at the same number of epochs. However, it is noteworthy that during our experiments, the computation time for KAN was frequently much longer compared to that of MLP (refer to the Appendix for more information).

\begin{figure*}[htbp]
\subfloat[\centering
  KAN
  \label{type3_kx_KAN}]
  {
    \includegraphics[width=0.45\linewidth]{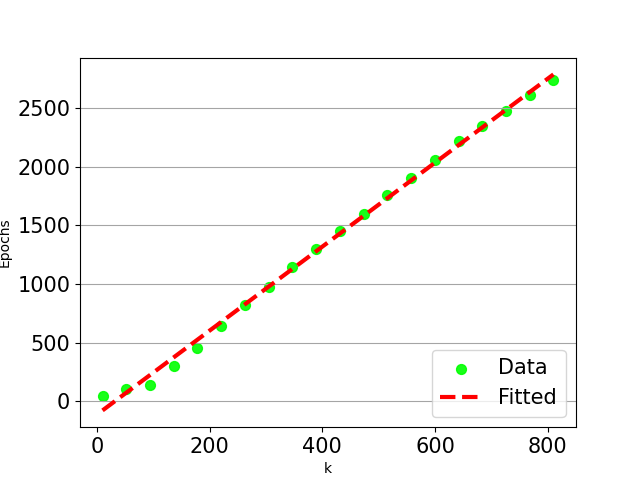}
  }
  \subfloat[\centering
  MLP
  \label{type3_kx_MLP}]
  {
    \includegraphics[width=0.45\linewidth]{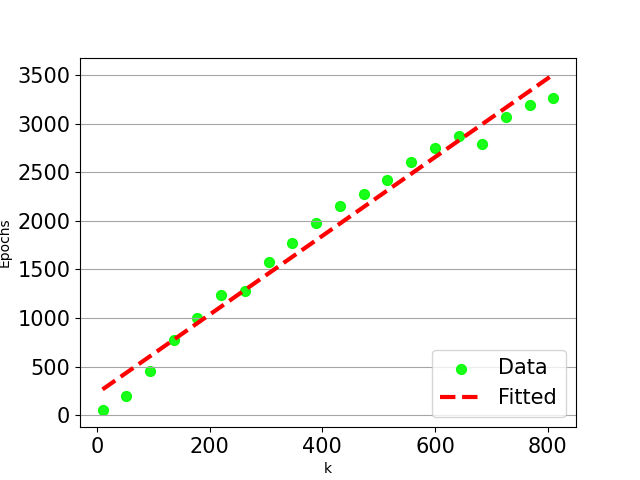}
  }

  \caption{Epochs minimization at error threshold $<1$ versus slope $k$: comparison between [1,79,1] MLP and [1,10,1] KAN (grid=3,k=3) in fitting function $f(x)=kx,x\in[0,1]$ }
  \label{type3_kx}
\end{figure*}

Fig \ref{type3_kx} sheds light on a potential reason for the below-par performance of the two networks in such function fitting tasks. Notably, within a given error tolerance limit, it has been noted that higher absolute values of the function's first derivative are associated with a greater number of epochs needed for convergence. Additionally, because of restrictions posed by the learning rate, increasing the number of epochs offers only marginal benefits in fitting accuracy, thus limiting the networks' overall performance.

\begin{table*}

\begin{minipage}[t]{0.48\textwidth}  
    \centering  
    \renewcommand\arraystretch{1.2}
    \caption{Time consumption of L-BFGS and Adam optimizers in fitting functions $f_7$ and $f_8$ using MLP and KAN} 
		\label{tab_t3_Opt}
		\begin{tabular}{|c|c|c|c|}
\hline

\textbf{Function} & \textbf{\textit{Network}}&\textbf{\textit{Optimizer}}& \textbf{\textit{Time(s)}}\\
\hline
$f_7$ &
MLP&L-BFGS&16.0677 \\
\hline
$f_7$ &
MLP&Adam&8.6278 \\
\hline
$f_7$ &
KAN&L-BFGS&901.4664\\
\hline
$f_7$ &
KAN&Adam&122.8072\\
\hline
$f_8$ &
MLP&L-BFGS&15.7724\\
\hline
$f_8$ &
MLP&Adam&8.4364\\
\hline
$f_8$ &
KAN&L-BFGS&941.1880\\
\hline
$f_8$ &
KAN&Adam&150.0152\\
\hline
\end{tabular}
\end{minipage}  
\renewcommand\arraystretch{1.2}
\hfill
\begin{minipage}[t]{0.48\textwidth}  
    \centering  
    \caption{Time consumption of L-BFGS and Adam optimizers in fitting functions $f_9$ and $f_{10}$ using MLP and KAN} 
		\label{tab_t4_Opt}
		\begin{tabular}{|c|c|c|c|}
\hline

\textbf{Function} & \textbf{\textit{Network}}&\textbf{\textit{Optimizer}}& \textbf{\textit{Time(s)}}\\
\hline
$f_9$ &
MLP&L-BFGS&14.7571 \\
\hline
$f_9$ &
MLP&Adam&7.8956 \\
\hline
$f_9$ &
KAN&L-BFGS&1365.7881\\
\hline
$f_9$ &
KAN&Adam&127.7374\\
\hline
$f_{10}$ &
MLP&L-BFGS&14.1521\\
\hline
$f_{10}$ &
MLP&Adam&8.5803\\
\hline
$f_{10}$ &
KAN&L-BFGS&557.7114\\
\hline
$f_{10}$ &
KAN&Adam&131.8761\\
\hline
\end{tabular}  
\end{minipage}
\end{table*}
\subsection{Functions with coherent oscillations}
A different kind of functional singularity, referred to as an 'coherent oscillatory singularity,' is illustrated by functions $f_9$ and $f_{10}$. These functions exhibit 'unreachable points' (for example, $x=0$ in the case of $f_9$) where, as the function gets closer to these points, its values oscillate more rapidly, crossing the x-axis an infinite number of times. At these 'unreachable points', not only does the absolute value of the first derivative lack a generalized limit, but the function values themselves cannot be represented within the extended real number system.

\begin{figure*}[htbp]
  \centering
  \subfloat[\centering
  $f_9$, Adam for MLP, L-BFGS for KAN
  \label{f9_trainingsamples}]
  {
    \includegraphics[width=0.47\linewidth]{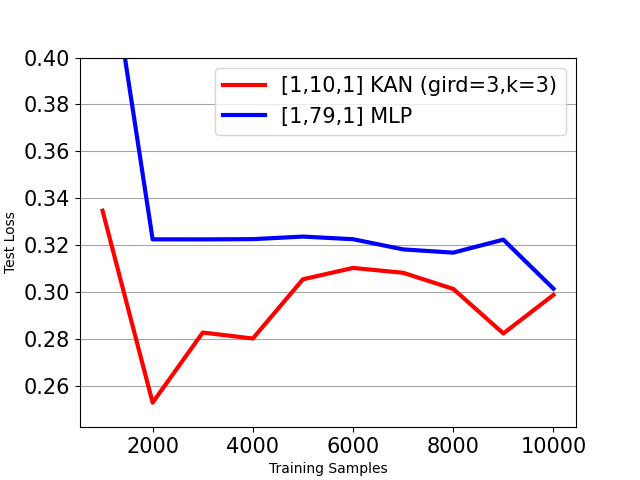}
  }
  \subfloat[\centering
  $f_{10}$, Adam for MLP and KAN
  \label{f10_trainingsamples}]
  {
    \includegraphics[width=0.47\linewidth]{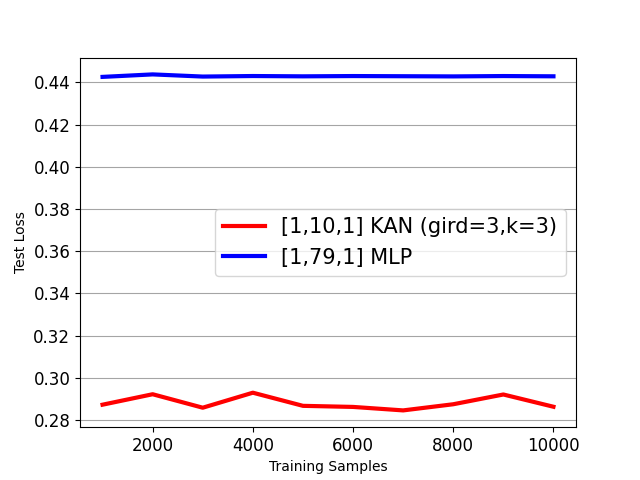}
  }

  \caption{Recover $f_9$ and $f_{10}$ independently using KAN and MLP with optimizer Adam or L-BFGS, 2000 Epochs}
  \label{type4_trainingsamples}
\end{figure*}

\begin{figure*}[htbp]
  \centering
  \subfloat[\centering
  MLP, $f_9$
  \label{f9_MLP_Opt}]
  {
    \includegraphics[width=0.23\linewidth]{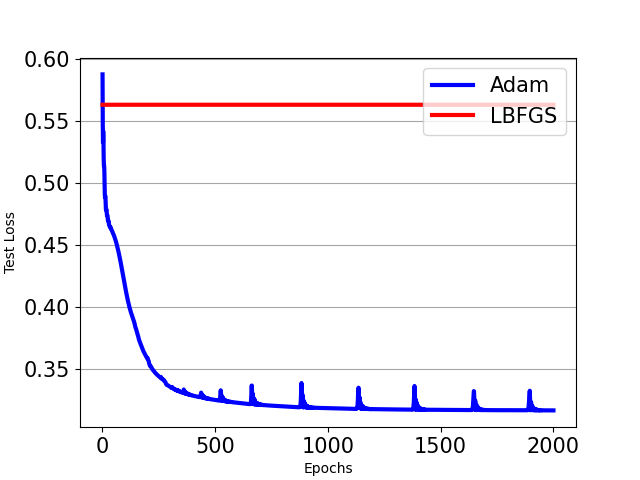}
  }
  \subfloat[\centering
  KAN, $f_9$
  \label{f9_KAN_Opt}]
  {
    \includegraphics[width=0.23\linewidth]{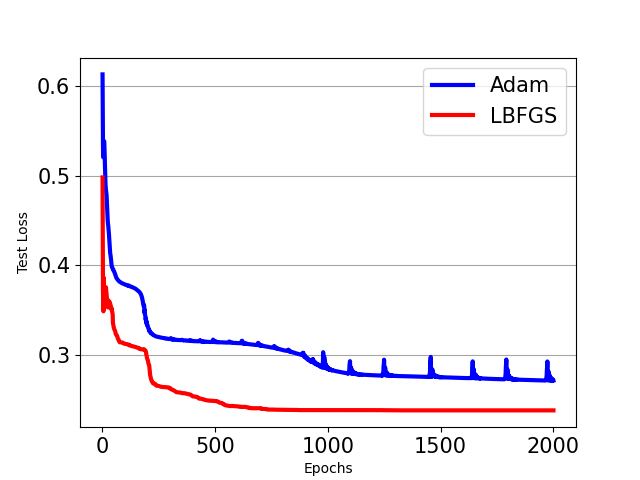}
  }
  \subfloat[\centering
  MLP, $f_{10}$
  \label{f10_MLP_Opt}]
  {
    \includegraphics[width=0.23\linewidth]{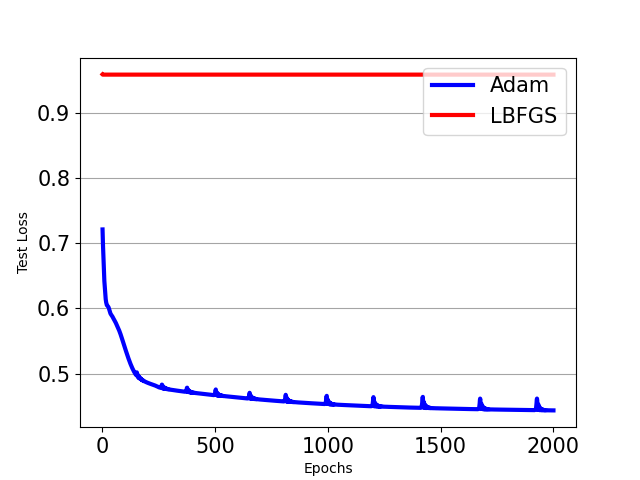}
  }
  \subfloat[\centering
  KAN, $f_{10}$
  \label{f10_KAN_Opt}]
  {
    \includegraphics[width=0.23\linewidth]{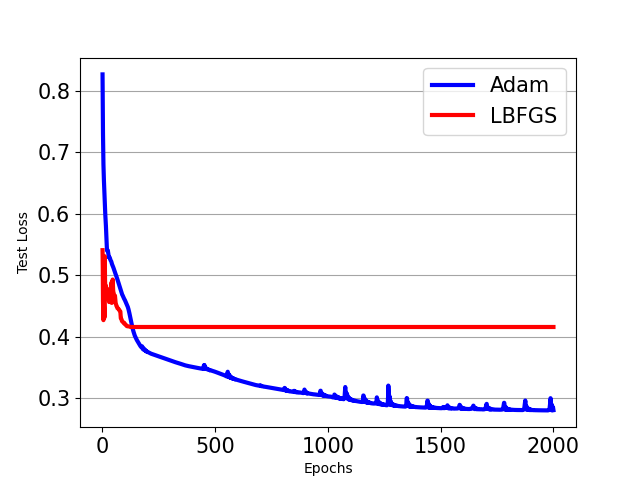}
  }

  \caption{The variation of test loss with the increasing number of epochs when recovering $f_9$ and $f_{10}$ independently using [1,10,1] KAN (grid=3,k=3) and [1,79,1] MLP with optimizer L-BFGS and Adam, learning rate=0.01}
  \label{type4_Opt}
\end{figure*}

In the experimental phase, taking a similar approach as described in section D, we initially explored the impacts of raising the sampling rate, the number of Epochs, and the selection of optimizer. As shown in Fig \ref{type4_trainingsamples} and \ref{type4_Opt}, an increase in the sampling rate did not markedly enhance fitting accuracy. Additionally, while increasing the number of Epochs was advantageous, its effectiveness diminished after surpassing a certain Epoch threshold, leading to a slower reduction in test loss.

In particular, within the KAN network framework, the optimizer L-BFGS outperformed Adam for function $f_9$, while for function $f_{10}$, Adam showed superior results. On the other hand, when fitting both functions with an MLP, Adam consistently performed better than L-BFGS.

\begin{figure*}[htbp]
  \centering
  \subfloat[\centering
  $f_9$, MLP (Adam)
  \label{f9_MLP}]
  {
    \includegraphics[width=0.23\linewidth]{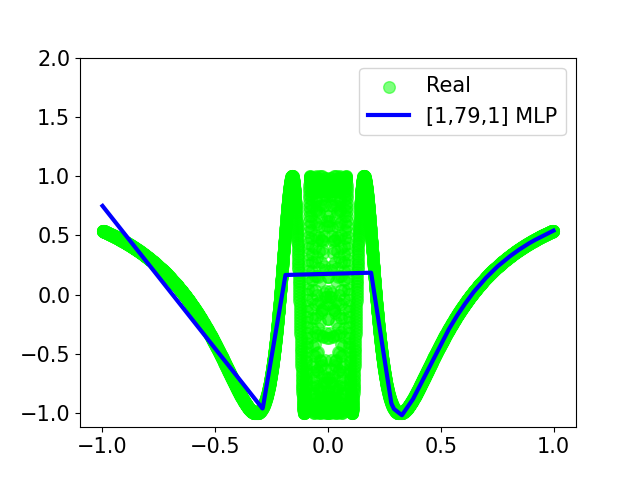}
  }
  \subfloat[\centering
  $f_9$, KAN (L-BFGS) 
  \label{f9_KAN}]
  {
    \includegraphics[width=0.23\linewidth]{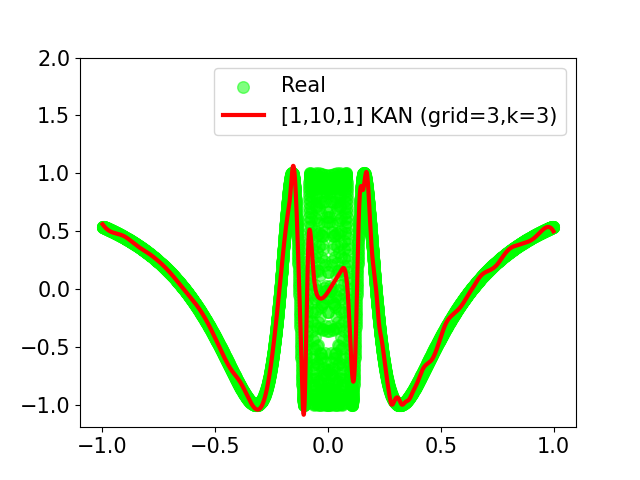}
  }
  \subfloat[\centering
  $f_{10}$, MLP (Adam)
  \label{f10_MLP}]
  {
    \includegraphics[width=0.23\linewidth]{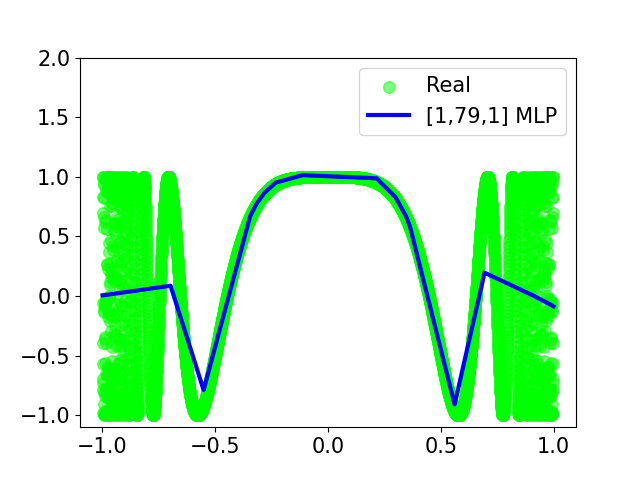}
  }
  \subfloat[\centering
  $f_{10}$, KAN (Adam)
  \label{f10_dif_epoch}]
  {
    \includegraphics[width=0.23\linewidth]{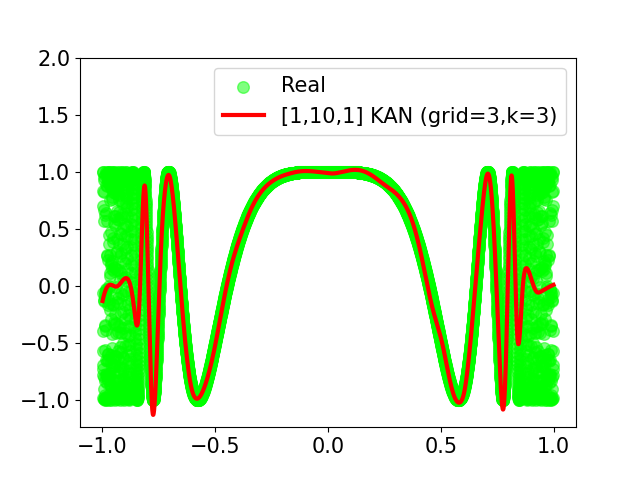}
  }

  \caption{Recover $f_9,(x\in [-0.999,0.999])$ and $f_{10},(x\in[-0.999,0.999])$ independently using [1,10,1] KAN (grid=3,k=3) and [1,79,1]MLP with same or different optimizers and epochs}
  \label{type4_fit}
\end{figure*}

In a similar manner, Table \ref{tab_t4_Opt} demonstrates that employing the L-BFGS optimizer during the fitting process usually resulted in an additional increase in computational time. Figure \ref{type4_fit} demonstrates that KAN consistently surpasses MLP when comparing performance over the same number of epochs.

\section{Noisy functions}\label{Cmp_noise}
In the following, we discuss the roles of noise functions. These functions have a wide range of applications, as noise can be added to any function, including those previously discussed. Thus, we introduce noise to these functions and proceed to evaluate the performance of MLP and KAN. According to the conclusions drawn in the preceding section, we will classify the functions into three categories: normal functions, functions with localized irregularities, and functions with severe discontinuities.

\subsection{Regular functions}

\begin{figure*}[htbp]
  \centering 
  \subfloat[\centering
  Various training samples, 
  {[1,1,1]} KAN (grid=3, k=3) and {[1,7,1]} MLP,
  $f_1$ with noise
  \label{f1_noise}]
  {
    \includegraphics[width=0.45\linewidth]{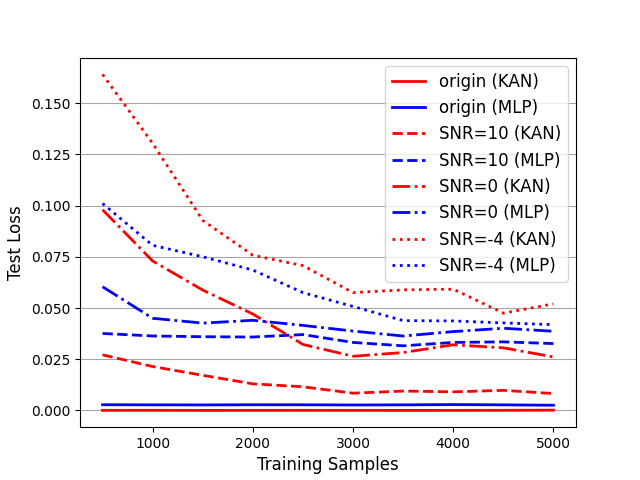}
  }
  \subfloat[\centering
  Various training samples,  
  {[1,5,1]} KAN (grid=3, k=3) and {[1,39,1]} MLP,
  $f_2$ with noise
  \label{f2_noise}]
  {
    \includegraphics[width=0.45\linewidth]{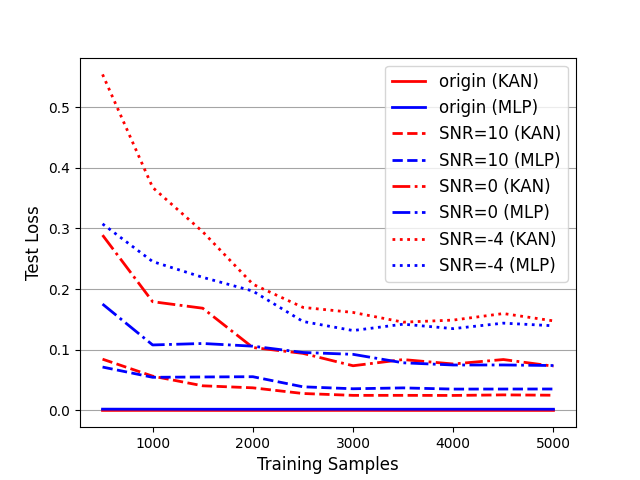}
  }

  \subfloat[\centering
  Original data, noisy data and predictions, 
  $f_1$ with noise, 
  training samples = 50 
  \label{f1_50_noise}]
  {
    \includegraphics[width=0.32\linewidth]{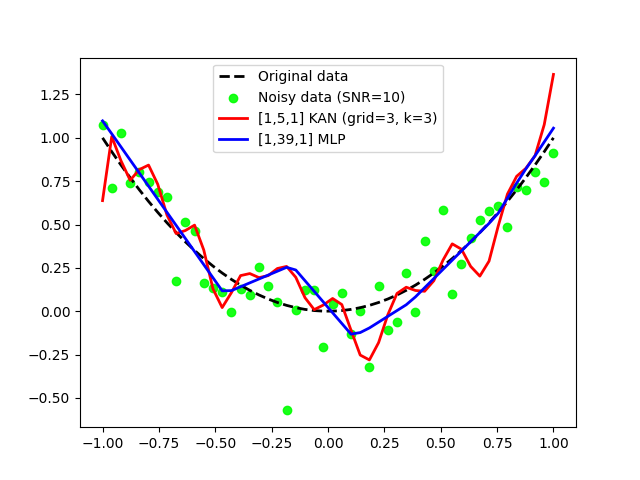}
  }
  \subfloat[\centering
  Original data, noisy data 
  and predictions of MLP, 
  $f_1$ with noise, 
  training samples = 5000 
  (500 plotted for clarity)
  \label{f1_5000_mlp_noise}]
  {
    \includegraphics[width=0.32\linewidth]{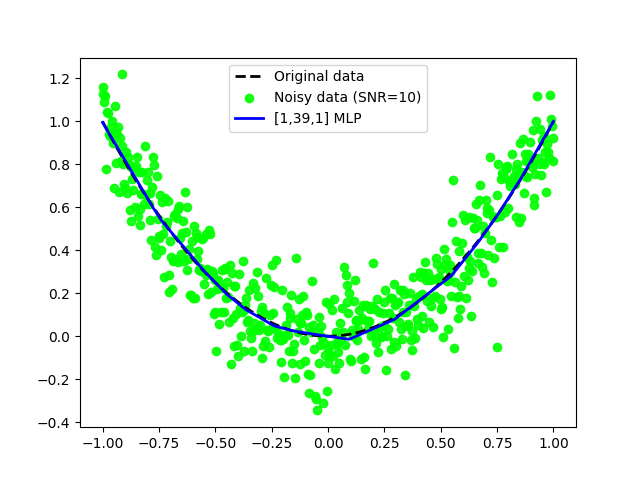}
  }
  \subfloat[\centering
  Original data, noisy data 
  and predictions of KAN, 
  $f_3$ with noise, 
  training samples = 5000 
  (500 plotted for clarity)
  \label{f1_5000_kan_noise}]
  {
    \includegraphics[width=0.32\linewidth]{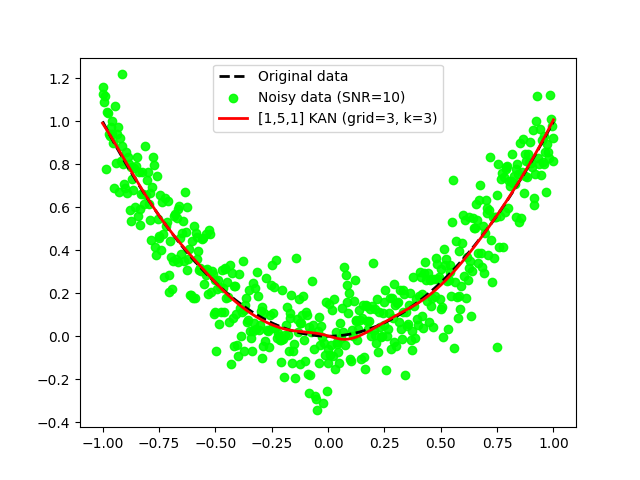}
  }

  \subfloat[\centering
  Original data, noisy data and predictions, 
  $f_2$ with noise, 
  training samples = 50 
  \label{f2_50_noise}]
  {
    \includegraphics[width=0.32\linewidth]{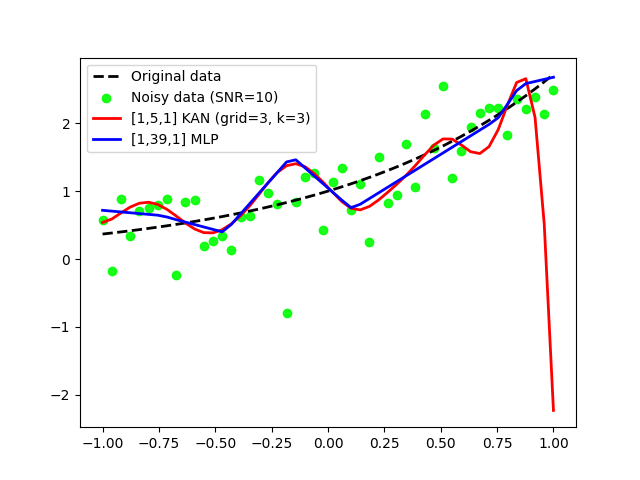}
  }
  \subfloat[\centering
  Original data, noisy data 
  and predictions of MLP, 
  $f_2$ with noise, 
  training samples = 5000 
  (500 plotted for clarity)
  \label{f2_5000_mlp_noise}]
  {
    \includegraphics[width=0.32\linewidth]{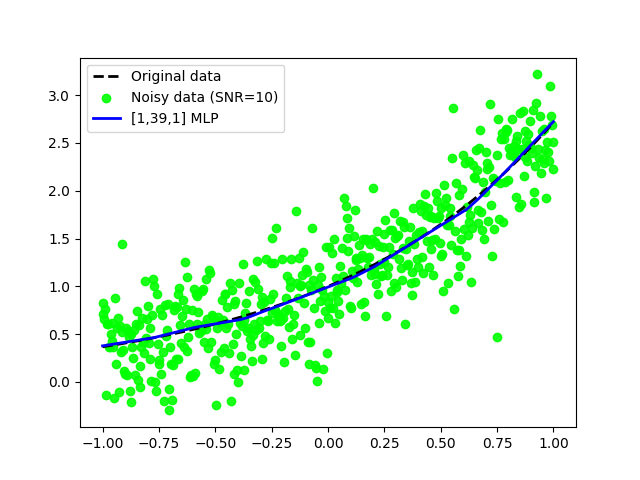}
  }
  \subfloat[\centering
  Original data, noisy data 
  and predictions of KAN, 
  $f_2$ with noise, 
  training samples = 5000 
  (500 plotted for clarity)
  \label{f2_5000_kan_noise}]
  {
    \includegraphics[width=0.32\linewidth]{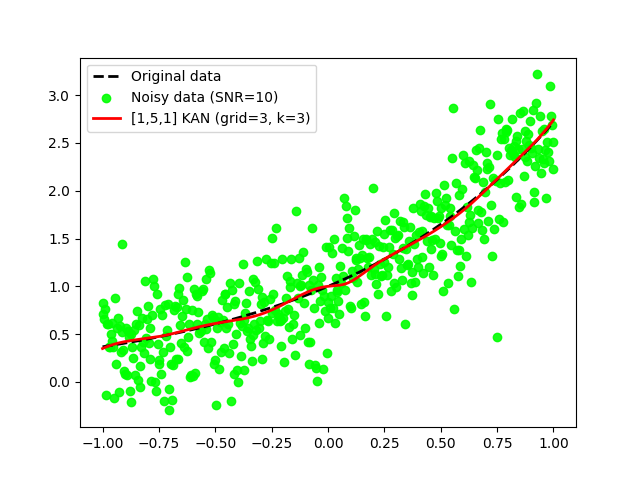}
  }

  \caption{Recover $f_1$ and $f_2$ with noise independently using KAN and MLP.}
  \label{type0_noise}
\end{figure*}

We introduce noise to functions exhibiting strong regularity, and subsequently fit these noisy data using KAN and MLP. The experimental outcomes are depicted in Fig \ref{type0_noise}. Our observations indicate that KAN achieves a lower test loss with low noise levels but performs worse under high noise conditions. When comparing the function fitting effect, the conclusion remains consistent: MLP shows better performance with minor noise interference, but KAN rapidly outperforms MLP as the training sample size increases.

\subsection{Adding noise to functions with  irregularities}

Noise is subsequently added to $f_3$, $f_4$, $f_5$, and $f_6$. The experimental findings are shown in Fig \ref{type12_noise}. It is clear that the noise nearly completely conceals the irregular characteristics of the functions. When compared to the original data, fitting noisy data is more difficult for neural networks. Although increasing the number of training samples greatly enhances the fitting performance, it still falls short of the noise-free level. For $f_3$ and $f_4$, the network can still capture some of the irregular features with a larger training sample. However, for $f_5$ and $f_6$, both KAN and MLP perform poorly. The networks still have difficulty identifying the jump discontinuities, even with an increased sample size.

\subsection{Irregular functions adding noise}

\begin{figure*}[htbp]
  \centering 
  \subfloat[\centering
  Various training samples,  
  {[1,10,1]} KAN (grid=3, k=3) and {[1,79,1]} MLP,
  $f_7$ with noise
  \label{f7_noise}]
  {
    \includegraphics[width=0.32\linewidth]{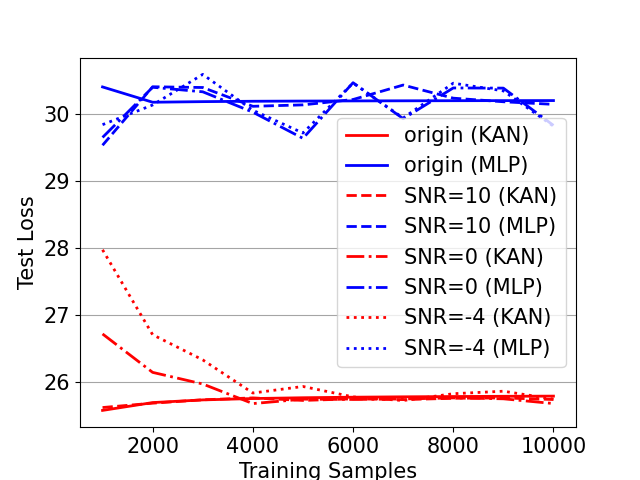}
  }
  \subfloat[\centering
  Original data, noisy data and predictions, 
  $f_7$ with noise, MLP 
  \label{f7_mlp_noise}]
  {
    \includegraphics[width=0.32\linewidth]{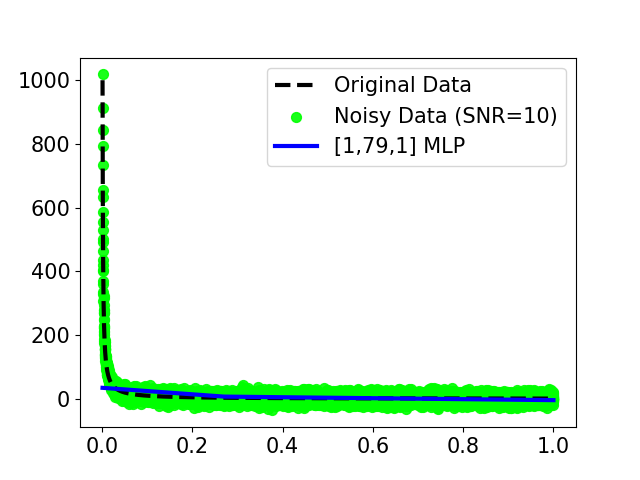}
  }
  \subfloat[\centering
  Original data, noisy data and predictions, 
  $f_7$ with noise, KAN
  \label{f7_kan_noise}]
  {
    \includegraphics[width=0.32\linewidth]{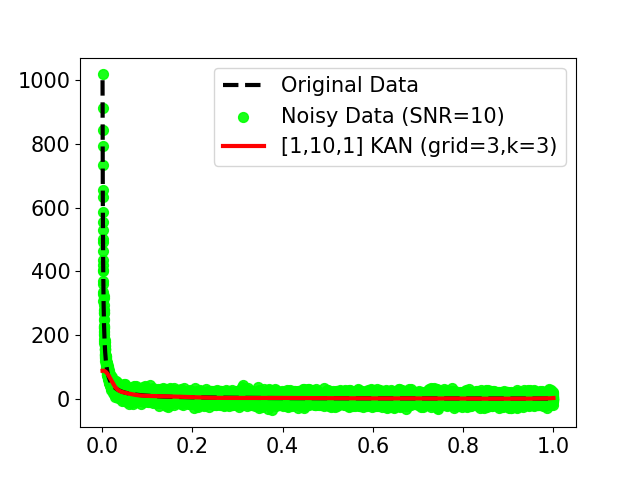}
  }

  \subfloat[\centering
  Various training samples,  
  {[1,10,1]} KAN (grid=3, k=3) and {[1,79,1]} MLP,
  $f_8$ with noise
  \label{f8_noise}]
  {
    \includegraphics[width=0.32\linewidth]{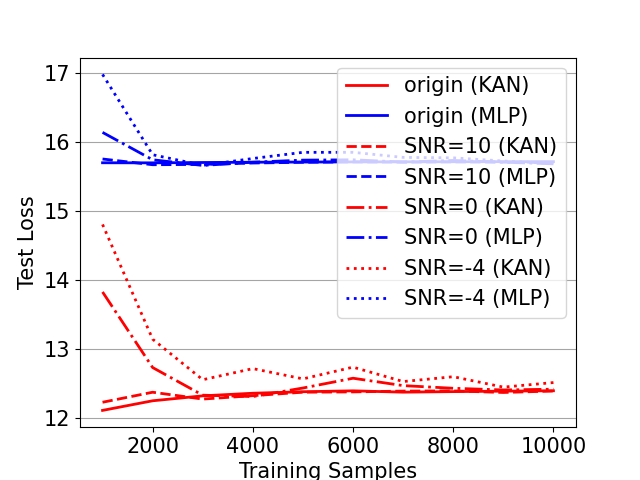}
  }
  \subfloat[\centering
  Original data, noisy data and predictions, 
  $f_8$ with noise, MLP
  \label{f8_mlp_noise}]
  {
    \includegraphics[width=0.32\linewidth]{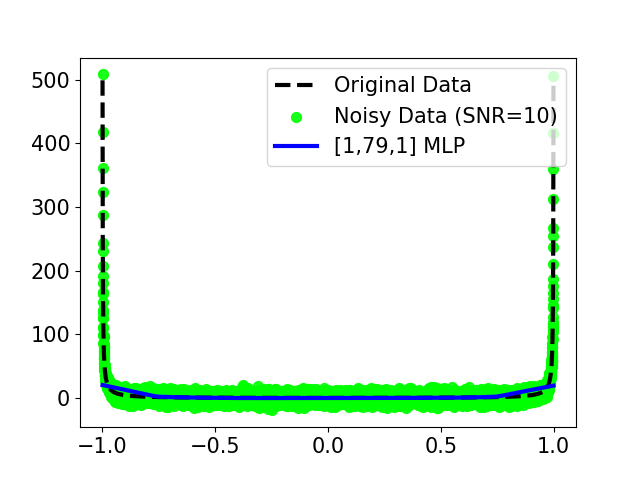}
  }
  \subfloat[\centering
  Original data, noisy data and predictions, 
  $f_8$ with noise, KAN
  \label{f8_kan_noise}]
  {
    \includegraphics[width=0.32\linewidth]{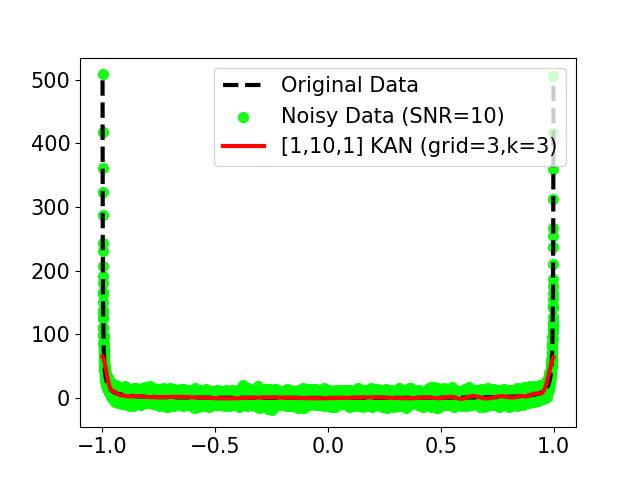}
  }

  \subfloat[\centering
  Various training samples,  
  {[1,10,1]} KAN (grid=3, k=3, Opt=L-BFGS) and {[1,79,1]} MLP (Opt=Adam),
  $f_9$ with noise
  \label{f9_noise}]
  {
    \includegraphics[width=0.32\linewidth]{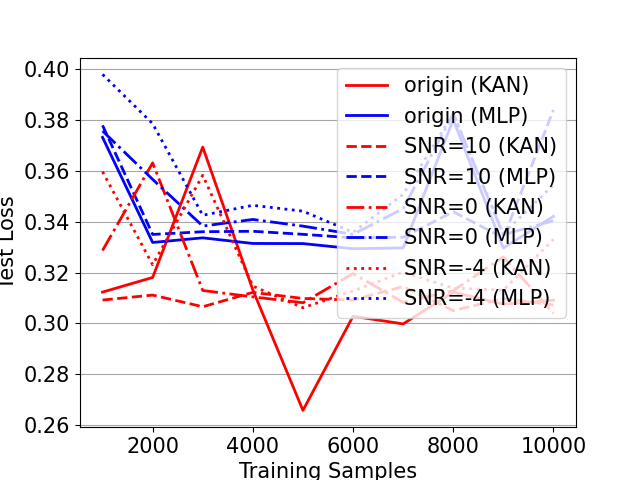}
  }
  \subfloat[\centering
  Original data, noisy data and predictions, 
  $f_9$ with noise, MLP
  \label{f9_mlp_noise}]
  {
    \includegraphics[width=0.32\linewidth]{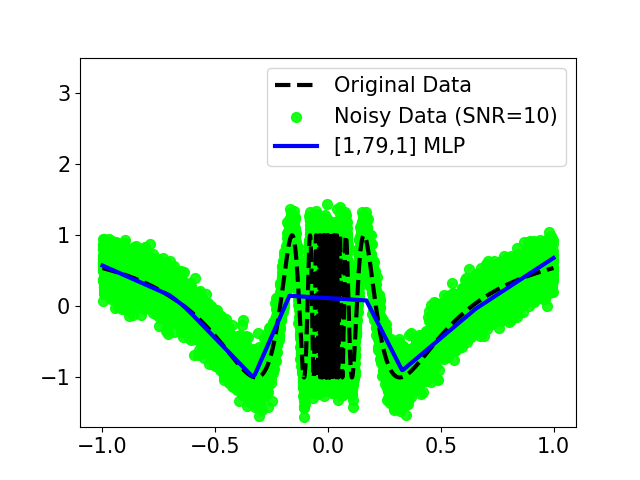}
  }
  \subfloat[\centering
  Original data, noisy data and predictions, 
  $f_9$ with noise, KAN
  \label{f9_kan_noise}]
  {
    \includegraphics[width=0.32\linewidth]{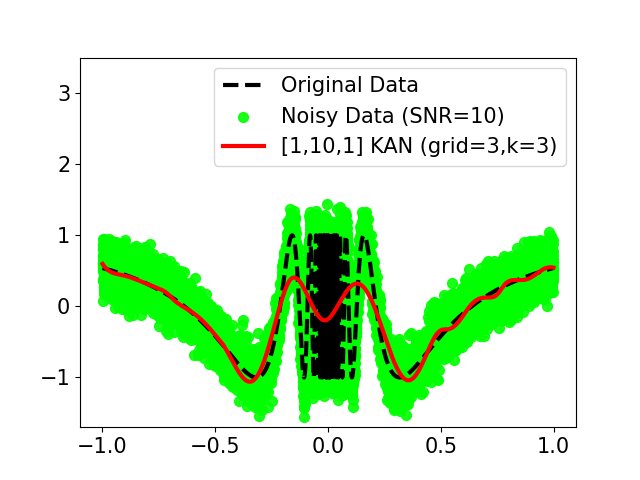}
  }

  \subfloat[\centering
  Various training samples,  
  {[1,10,1]} KAN (grid=3, k=3) and {[1,79,1]} MLP,
  $f_{10}$ with noise
  \label{f10_noise}]
  {
    \includegraphics[width=0.32\linewidth]{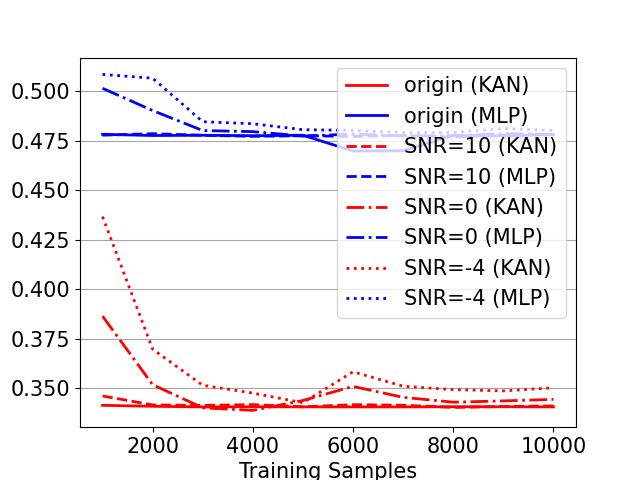}
  }
  \subfloat[\centering
  Original data, noisy data and predictions, 
  $f_{10}$ with noise, MLP
  \label{f10_mlp_noise}]
  {
    \includegraphics[width=0.32\linewidth]{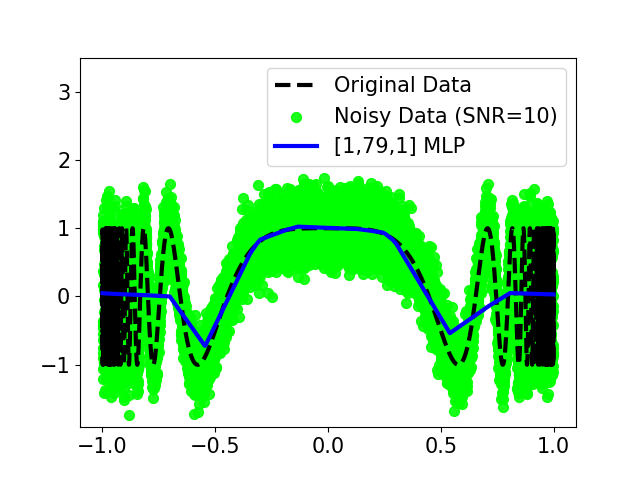}
  }
  \subfloat[\centering
  Original data, noisy data and predictions, 
  $f_{10}$ with noise, KAN
  \label{f10_10000_noise}]
  {
    \includegraphics[width=0.32\linewidth]{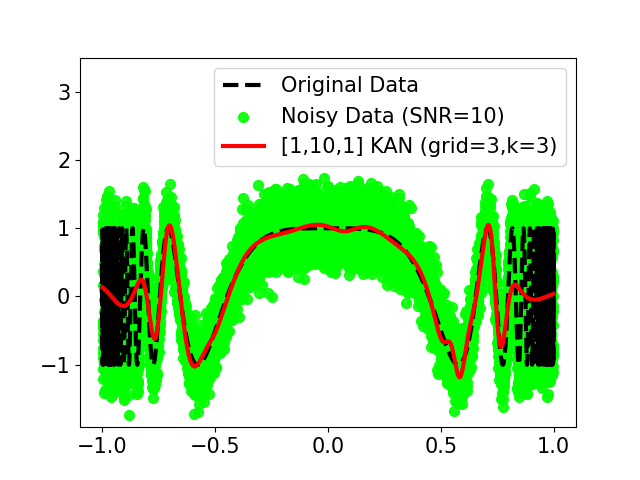}
  }
  
  \caption{Recover noisy functions with singularities or coherent oscillations independently using KAN and MLP.}
  \label{type34_noise}
\end{figure*}

In Fig \ref{type34_noise}, it becomes clear that the KAN fitting performance greatly surpasses that of MLP when noise is added to functions with singularities or coherent oscillation. Interestingly, in these extreme and unusual function scenarios, the impact of adding noise to the fitting process is minimal, to the point that fitting with noisy data produces results that are either on par with or better than those derived from the original, noise-free data. This minimal influence of noise highlights the ineffectiveness of strategies relying solely on increased sampling rates in such instances, as they fail to exploit the subtle benefits that noise brings.

\section{Conclusion}\label{Conclusion}
In this study, we evaluate the effectiveness of KAN and MLP in approximating irregular or noisy functions. Our analysis concentrates on two main factors: the relative performance of KAN and MLP in fitting functions with different types according to  regularity, and their ability to handle noise during the fitting process.

At first, in order to thoroughly evaluate the performance of the two networks, we designed three sets of comparative experiments. These focus on testing their effectiveness in fitting different functions under varied conditions: changing sampling rates, varying numbers of epochs, and, particularly for functions $f_7-f_{10}$, examining the effect of different optimizers (exclusively to identify the optimal optimizer).

Secondly, as identified in \cite{our_2024kan_With_noise} and additionally explored in this study, raising the sampling rate is a potent method to enhance the fitting performance of functions $f_1-f_6$. Particularly, this strategy shows greater advantages when handling noisy data versus clean data. Nevertheless, the improvement in the fitting accuracy for functions with low regularity ($f_7-f_{10}$) is minimal, irrespective of the presence of noise.

Thirdly, we also compared the fitting performance under varying Epochs from two distinct perspectives: convergence speed and stabilized test loss. KAN exhibits a faster convergence rate than MLP across all tested functions. However, MLP outperforms KAN on test functions $f_3-f_6$ on stabilized test loss.

Fourthly, via experimental analysis (fitting $f_7-f_{10}$), it was observed that Adam exceeded L-BFGS in performance for both networks in every instance, except for function $f_9$. Notably, when fitting function $f_9$ with the KAN, L-BFGS demonstrated better results than Adam.

At last, when dealing with noisy functions, KAN exhibits superior performance over MLP for regular functions or irregular functions. Conversely, for functions that with jump discontinuities or singularities, MLP outperforms KAN.

\section*{Acknowledgment}
The authors wish to extend their sincere appreciation to Dr.~Aijun Zhang for his enlightening discussions about the negative effects of noise on the KANs network and for offering numerous invaluable suggestions for this paper.

\bibliographystyle{IEEEtran}

\bibliography{KAN-2.bib}

\begin{thebibliography}{10}
\providecommand{\url}[1]{#1}
\csname url@samestyle\endcsname
\providecommand{\newblock}{\relax}
\providecommand{\bibinfo}[2]{#2}
\providecommand{\BIBentrySTDinterwordspacing}{\spaceskip=0pt\relax}
\providecommand{\BIBentryALTinterwordstretchfactor}{4}
\providecommand{\BIBentryALTinterwordspacing}{\spaceskip=\fontdimen2\font plus
\BIBentryALTinterwordstretchfactor\fontdimen3\font minus \fontdimen4\font\relax}
\providecommand{\BIBforeignlanguage}[2]{{%
\expandafter\ifx\csname l@#1\endcsname\relax
\typeout{** WARNING: IEEEtran.bst: No hyphenation pattern has been}%
\typeout{** loaded for the language `#1'. Using the pattern for}%
\typeout{** the default language instead.}%
\else
\language=\csname l@#1\endcsname
\fi
#2}}
\providecommand{\BIBdecl}{\relax}
\BIBdecl

\bibitem{liu2024kan}
Z.~Liu, Y.~Wang, S.~Vaidya, F.~Ruehle, J.~Halverson, M.~Soljačić, T.~Y. Hou, and M.~Tegmark, ``Kan: Kolmogorov-arnold networks,'' 2024.

\bibitem{our_2024kan_With_noise}
H.~Shen, C.~Zeng, J.~Wang, and Q.~Wang, ``Reduced effectiveness of kolmogorov-arnold networks on functions with noise,'' 2024.

\bibitem{Kolmogorov1956-2}
A.~N. Kolmogorov, ``On the representation of continuous functions of several variables as superpositions of continuous functions of a smaller number of variables,'' \emph{Doklady Akademii Nauk SSSR}, vol. 108, no.~2, pp. 179--182, 1956.

\bibitem{Daubechies}
\BIBentryALTinterwordspacing
I.~Daubechies, \emph{Ten Lectures on Wavelets}.\hskip 1em plus 0.5em minus 0.4em\relax Society for Industrial and Applied Mathematics, 1992. [Online]. Available: \url{https://epubs.siam.org/doi/abs/10.1137/1.9781611970104}
\BIBentrySTDinterwordspacing

\bibitem{Zhang2024}
A.~Zhang, ``Kans can't deal with noise,'' 2024, \url{https://github.com/SelfExplainML/PiML-Toolbox/blob/main/docs/Workshop/KANs\_Can't\_Deal\_with\_Noise.ipynb}.

\bibitem{kernel2024}
A.~A. Gomez, A.~S. Neto, and J.~Zubelli, ``Diffusion representation for asymmetric kernels,'' 2024.

\bibitem{Kolmogorov1957}
A.~N. Kolmogorov, ``\BIBforeignlanguage{Russian}{On the representation of continuous functions of several variables by superpositions of continuous functions of one variable and addition},'' \emph{\BIBforeignlanguage{Russian}{Doklady Akademii Nauk SSSR}}, vol. 114, pp. 953--956, 1957.

\bibitem{Arnold1957}
V.~I. Arnold, ``\BIBforeignlanguage{Russian}{On functions of three variables},'' \emph{\BIBforeignlanguage{Russian}{Doklady Akademii Nauk SSSR}}, vol. 114, pp. 679--681, 1957.

\bibitem{Papoulis1977SignalA}
\BIBentryALTinterwordspacing
A.~Papoulis, ``Signal analysis,'' 1977. [Online]. Available: \url{https://api.semanticscholar.org/CorpusID:115354299}
\BIBentrySTDinterwordspacing

\bibitem{Braun2009}
\BIBentryALTinterwordspacing
J.~Braun and M.~Griebel, ``On a constructive proof of kolmogorov’s superposition theorem,'' \emph{Constructive Approximation}, vol.~30, pp. 653--675, 2009. [Online]. Available: \url{https://doi.org/10.1007/s00365-009-9054-2}
\BIBentrySTDinterwordspacing

\bibitem{285644}
S.~Hein and A.~Zakhor, ``Reconstruction of oversampled band-limited signals from sigma/delta encoded binary sequences,'' \emph{IEEE Transactions on Signal Processing}, vol.~42, no.~4, pp. 799--811, 1994.

\bibitem{1468473}
I.~Maravic and M.~Vetterli, ``Sampling and reconstruction of signals with finite rate of innovation in the presence of noise,'' \emph{IEEE Transactions on Signal Processing}, vol.~53, no.~8, pp. 2788--2805, 2005.

\bibitem{oversampling}
J.~Zhang, T.~Wang, W.~Ng, and W.~Pedrycz, ``Perturbation-based oversampling technique for imbalanced classification problems,'' \emph{International Journal of Machine Learning and Cybernetics}, vol.~14, pp. 773--787, 2023.

\bibitem{2024arXiv240508790V}
C.~J. {Vaca-Rubio}, L.~{Blanco}, R.~{Pereira}, and M.~{Caus}, ``{Kolmogorov-Arnold Networks (KANs) for Time Series Analysis},'' \emph{arXiv e-prints}, p. arXiv:2405.08790, May 2024.

\bibitem{2024arXiv240602496X}
K.~{Xu}, L.~{Chen}, and S.~{Wang}, ``{Kolmogorov-Arnold Networks for Time Series: Bridging Predictive Power and Interpretability},'' \emph{arXiv e-prints}, p. arXiv:2406.02496, Jun. 2024.

\bibitem{2024arXiv240704192K}
B.~C. {Koenig}, S.~{Kim}, and S.~{Deng}, ``{KAN-ODEs: Kolmogorov-Arnold Network Ordinary Differential Equations for Learning Dynamical Systems and Hidden Physics},'' \emph{arXiv e-prints}, p. arXiv:2407.04192, Jul. 2024.

\bibitem{2024arXiv240611045W}
Y.~{Wang}, J.~{Sun}, J.~{Bai}, C.~{Anitescu}, M.~{Sadegh Eshaghi}, X.~{Zhuang}, T.~{Rabczuk}, and Y.~{Liu}, ``{Kolmogorov Arnold Informed neural network: A physics-informed deep learning framework for solving PDEs based on Kolmogorov Arnold Networks},'' \emph{arXiv e-prints}, p. arXiv:2406.11045, Jun. 2024.

\bibitem{2024arXiv240615719J}
A.~{Jamali}, S.~K. {Roy}, D.~{Hong}, B.~{Lu}, and P.~{Ghamisi}, ``{How to Learn More? Exploring Kolmogorov-Arnold Networks for Hyperspectral Image Classification},'' \emph{arXiv e-prints}, p. arXiv:2406.15719, Jun. 2024.

\bibitem{2024arXiv240507488P}
Y.~{Peng}, M.~{He}, F.~{Hu}, Z.~{Mao}, X.~{Huang}, and J.~{Ding}, ``{Predictive Modeling of Flexible EHD Pumps using Kolmogorov-Arnold Networks},'' \emph{arXiv e-prints}, p. arXiv:2405.07488, May 2024.

\bibitem{2024arXiv240609087A}
B.~{Azam} and N.~{Akhtar}, ``{Suitability of KANs for Computer Vision: A preliminary investigation},'' \emph{arXiv e-prints}, p. arXiv:2406.09087, Jun. 2024.

\bibitem{2024arXiv240618380B}
R.~{Bresson}, G.~{Nikolentzos}, G.~{Panagopoulos}, M.~{Chatzianastasis}, J.~{Pang}, and M.~{Vazirgiannis}, ``{KAGNNs: Kolmogorov-Arnold Networks meet Graph Learning},'' \emph{arXiv e-prints}, p. arXiv:2406.18380, Jun. 2024.

\bibitem{2024arXiv240613597Z}
F.~{Zhang} and X.~{Zhang}, ``{GraphKAN: Enhancing Feature Extraction with Graph Kolmogorov Arnold Networks},'' \emph{arXiv e-prints}, p. arXiv:2406.13597, Jun. 2024.

\bibitem{2024arXiv240602918L}
C.~{Li}, X.~{Liu}, W.~{Li}, C.~{Wang}, H.~{Liu}, and Y.~{Yuan}, ``{U-KAN Makes Strong Backbone for Medical Image Segmentation and Generation},'' \emph{arXiv e-prints}, p. arXiv:2406.02918, Jun. 2024.

\bibitem{2024arXiv240507200S}
S.~Sidharth, A.~Keerthana, R.~Gokul, and K.~Anas, ``{Chebyshev Polynomial-Based Kolmogorov-Arnold Networks: An Efficient Architecture for Nonlinear Function Approximation},'' \emph{arXiv e-prints}, p. arXiv:2405.07200, May 2024.

\bibitem{2024arXiv240606470K}
M.~{Kiamari}, M.~{Kiamari}, and B.~{Krishnamachari}, ``{GKAN: Graph Kolmogorov-Arnold Networks},'' \emph{arXiv e-prints}, p. arXiv:2406.06470, Jun. 2024.

\bibitem{2024arXiv240507344G}
R.~{Genet} and H.~{Inzirillo}, ``{TKAN: Temporal Kolmogorov-Arnold Networks},'' \emph{arXiv e-prints}, p. arXiv:2405.07344, May 2024.

\bibitem{2024arXiv240512832B}
Z.~{Bozorgasl} and H.~{Chen}, ``{Wav-KAN: Wavelet Kolmogorov-Arnold Networks},'' \emph{arXiv e-prints}, p. arXiv:2405.12832, May 2024.

\bibitem{2024arXiv240613155D}
A.~{Dylan Bodner}, A.~{Santiago Tepsich}, J.~{Natan Spolski}, and S.~{Pourteau}, ``{Convolutional Kolmogorov-Arnold Networks},'' \emph{arXiv e-prints}, p. arXiv:2406.13155, Jun. 2024.

\bibitem{2024arXiv240607456A}
A.~{Afzal Aghaei}, ``{fKAN: Fractional Kolmogorov-Arnold Networks with trainable Jacobi basis functions},'' \emph{arXiv e-prints}, p. arXiv:2406.07456, Jun. 2024.

\bibitem{2024arXiv240602075Q}
Q.~{Qiu}, T.~{Zhu}, H.~{Gong}, L.~{Chen}, and H.~{Ning}, ``{ReLU-KAN: New Kolmogorov-Arnold Networks that Only Need Matrix Addition, Dot Multiplication, and ReLU},'' \emph{arXiv e-prints}, p. arXiv:2406.02075, Jun. 2024.

\bibitem{Thao-Vetterli}
N.~Thao and M.~Vetterli, ``Reduction of the mse in r-times oversampled a/d conversion o$(1/r)$ to o$(1/r^2)$,'' \emph{IEEE Transactions on Signal Processing}, vol.~42, no.~1, pp. 200--203, 1994.

\bibitem{coifman2006diffusion}
R.~R. Coifman and S.~Lafon, ``Diffusion maps,'' \emph{Applied and Computational Harmonic Analysis}, vol.~21, no.~1, pp. 5--30, 2006.

\bibitem{wang2024blackboxclarityaipowered}
\BIBentryALTinterwordspacing
X.~Wang, Y.~Li, Y.~Li, and G.~Kish, ``From black box to clarity: Ai-powered smart grid optimization with kolmogorov-arnold networks,'' 2024. [Online]. Available: \url{https://arxiv.org/abs/2408.04063}
\BIBentrySTDinterwordspacing

\bibitem{limitationclassification}
a.~Tran, T.~Le, D.~Tran, P.~Hoai~Luan, D.~Le~Vu~Trung, T.~Vu, V.~Nguyen, and Y.~Nakashima, ``Exploring the limitations of kolmogorov-arnold networks in classification: Insights to software training and hardware implementation,'' 07 2024.

\bibitem{yu2024kanmlpfairercomparison}
\BIBentryALTinterwordspacing
R.~Yu, W.~Yu, and X.~Wang, ``Kan or mlp: A fairer comparison,'' 2024. [Online]. Available: \url{https://arxiv.org/abs/2407.16674}
\BIBentrySTDinterwordspacing

\bibitem{2024arXiv240616026W}
J.~{Wang}, P.~{Cai}, Z.~{Wang}, H.~{Zhang}, and J.~{Huang}, ``{CEST-KAN: Kolmogorov-Arnold Networks for CEST MRI Data Analysis},'' \emph{arXiv e-prints}, p. arXiv:2406.16026, Jun. 2024.

\bibitem{2024arXiv240600600C}
M.~{Cheon}, ``{Kolmogorov-Arnold Network for Satellite Image Classification in Remote Sensing},'' \emph{arXiv e-prints}, p. arXiv:2406.00600, Jun. 2024.

\bibitem{2024arXiv240704819Z}
J.~{Zhang}, ``{RPN: Reconciled Polynomial Network Towards Unifying PGMs, Kernel SVMs, MLP and KAN},'' \emph{arXiv e-prints}, p. arXiv:2407.04819, Jul. 2024.

\bibitem{2024arXiv240617630K}
A.~{Kundu}, A.~{Sarkar}, and A.~{Sadhu}, ``{KANQAS: Kolmogorov-Arnold Network for Quantum Architecture Search},'' \emph{arXiv e-prints}, p. arXiv:2406.17630, Jun. 2024.

\bibitem{2024arXiv240607869T}
S.~{Teymoor Seydi}, ``{Unveiling the Power of Wavelets: A Wavelet-based Kolmogorov-Arnold Network for Hyperspectral Image Classification},'' \emph{arXiv e-prints}, p. arXiv:2406.07869, Jun. 2024.

\bibitem{2024arXiv240800273T}
T.~{Tang}, Y.~{Chen}, and H.~{Shu}, ``{3D U-KAN Implementation for Multi-modal MRI Brain Tumor Segmentation},'' \emph{arXiv e-prints}, p. arXiv:2408.00273, Aug. 2024.

\bibitem{2024arXiv240802694W}
T.~{Wang} and S.~{Singh}, ``{KAN based Autoencoders for Factor Models},'' \emph{arXiv e-prints}, p. arXiv:2408.02694, Aug. 2024.

\bibitem{2024arXiv240801018L}
R.~{Li}, ``{GNN-MolKAN: Harnessing the Power of KAN to Advance Molecular Representation Learning with GNNs},'' \emph{arXiv e-prints}, p. arXiv:2408.01018, Aug. 2024.

\bibitem{2024arXiv240802088L}
Z.~{Lai}, C.~{Liu}, S.~{Sheng}, and Z.~{Zhang}, ``{KAN-RCBEVDepth: A multi-modal fusion algorithm in object detection for autonomous driving},'' \emph{arXiv e-prints}, p. arXiv:2408.02088, Aug. 2024.

\bibitem{2024arXiv240710347L}
A.~{Lawan}, J.~{Pu}, H.~{Yunusa}, A.~{Umar}, and M.~{Lawan}, ``{MambaForGCN: Enhancing Long-Range Dependency with State Space Model and Kolmogorov-Arnold Networks for Aspect-Based Sentiment Analysis},'' \emph{arXiv e-prints}, p. arXiv:2407.10347, Jul. 2024.

\bibitem{2024arXiv240713044G}
M.~{Ghaith Altarabichi}, ``{DropKAN: Regularizing KANs by masking post-activations},'' \emph{arXiv e-prints}, p. arXiv:2407.13044, Jul. 2024.

\bibitem{2024arXiv240716268I}
A.~{Igali} and P.~{Shamoi}, ``{Image Classification using Fuzzy Pooling in Convolutional Kolmogorov-Arnold Networks},'' \emph{arXiv e-prints}, p. arXiv:2407.16268, Jul. 2024.

\bibitem{2024arXiv240721176Z}
S.~{Zinage}, S.~{Mondal}, and S.~{Sarkar}, ``{DKL-KAN: Scalable Deep Kernel Learning using Kolmogorov-Arnold Networks},'' \emph{arXiv e-prints}, p. arXiv:2407.21176, Jul. 2024.

\bibitem{2024arXiv240720265L}
X.~{Li}, Z.~{Feng}, Y.~{Chen}, W.~{Dai}, Z.~{He}, Y.~{Zhou}, and S.~{Jiao}, ``{COEFF-KANs: A Paradigm to Address the Electrolyte Field with KANs},'' \emph{arXiv e-prints}, p. arXiv:2407.20265, Jul. 2024.

\bibitem{2024arXiv240706560D}
F.~{Dong}, ``{TCKIN: A Novel Integrated Network Model for Predicting Mortality Risk in Sepsis Patients},'' \emph{arXiv e-prints}, p. arXiv:2407.06560, Jul. 2024.

\bibitem{2024arXiv240609931C}
Y.~{Chen}, Z.~{Zhu}, S.~{Zhu}, L.~{Qiu}, B.~{Zou}, F.~{Jia}, Y.~{Zhu}, C.~{Zhang}, Z.~{Fang}, F.~{Qin}, J.~{Fan}, C.~{Wang}, Y.~{Gao}, and G.~{Yu}, ``{SCKansformer: Fine-Grained Classification of Bone Marrow Cells via Kansformer Backbone and Hierarchical Attention Mechanisms},'' \emph{arXiv e-prints}, p. arXiv:2406.09931, Jun. 2024.

\end{thebibliography}

\section*{Appendix}
Given the considerably extended computation times for KAN compared to MLP when fitting functions exhibiting singularities or coherent oscillations, as detailed in Section 3, we carried out further experiments to explore this matter. Nonetheless, owing to the persistent effects of coding-level optimizations, these findings do not serve as conclusive evidence. Therefore, they are included in the Appendix for documentation and further examination.

\vskip 0.4cm
\begin{minipage}{\columnwidth}  
\renewcommand\arraystretch{1.2}
\centering 
\captionof{table}{Performance of MLP and KAN in recovering $f_7$, $f_8$, $f_9$ and $f_{10}$} 
\label{tab_fit}
\centering
\begin{threeparttable}
\begin{tabular}{|c|c|c|c|c|}
\hline

\textbf{Function} & \textbf{\textit{Network}}&\textbf{\textit{Epochs}}& \textbf{\textit{Time(s)}}
&\makecell{\textbf{\textit{Test Loss}}\\ \textbf{\textit{(RMSE)}}}\\
\hline
$f_7$ &
MLP&2000&9.4114&22.17 \\
\hline
$f_7$ &
KAN&2000&124.5332&3.04 \\
\hline
$f_8$ &
MLP&2000&10.2214&11.20\\
\hline
$f_8$ &
KAN&2000&128.6405&4.72\\
\hline
$f_7$ &
MLP&25000&135.06336&0.95\\
\hline
$f_7$ &
KAN\tnote{*}&2000&149.6529&3.04\\
\hline
$f_8$ &
MLP&25000&103.8949&2.71\\
\hline
$f_8$ &
KAN\tnote{*}&2000&121.0088&4.72\\
\hline
$f_9$ &
MLP&2000&8.6413&0.32 \\
\hline
$f_9$ &
KAN&2000&993.7106&0.30 \\
\hline
$f_{10}$ &
MLP&2000&10.0180&0.42\\
\hline
$f_{10}$ &
KAN&2000&134.2680&0.28\\
\hline
$f_9$ &
MLP&220000&858.6642&0.23\\
\hline
$f_9$ &
KAN\tnote{*}&2000&1036.1291&0.30\\
\hline
$f_{10}$ &
MLP&22000&112.6193&0.37\\
\hline
$f_{10}$ &
KAN\tnote{*}&2000&118.9079&0.28\\
\hline
\end{tabular}
\begin{tablenotes}  
\footnotesize  
\item[*] denotes replicated experiments. For the sake of ensuring full alignment of the data with original results, they are explicitly presented.
\end{tablenotes}

\end{threeparttable} 
\end{minipage}  
\vskip 0.4cm

As shown in Table \ref{tab_fit}, by extending the number of epochs for training the MLP until its computation time matches that of the KAN, the MLP demonstrates better performance than the KAN in fitting the functions $f_7$ and $f_8$. However, the KAN outperforms the MLP for functions $f_9$ and $f_{10}$.

\begin{figure*}[htbp]
  \centering 
  \subfloat[\centering
  Various training samples, 
  {[1,1,1]} KAN (grid=3, k=3) and {[1,7,1]} MLP,
  $f_3$ with noise
  \label{f3_noise}]
  {
    \includegraphics[width=0.45\linewidth]{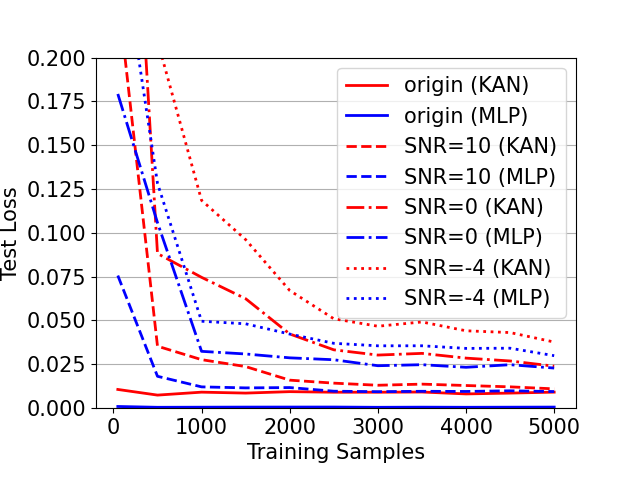}
  }
  \subfloat[\centering
  Various training samples, 
  {[1,5,1]} KAN (grid=3, k=3) and {[1,39,1]} MLP,
  $f_4$ with noise
  \label{f4_noise}]
  {
    \includegraphics[width=0.45\linewidth]{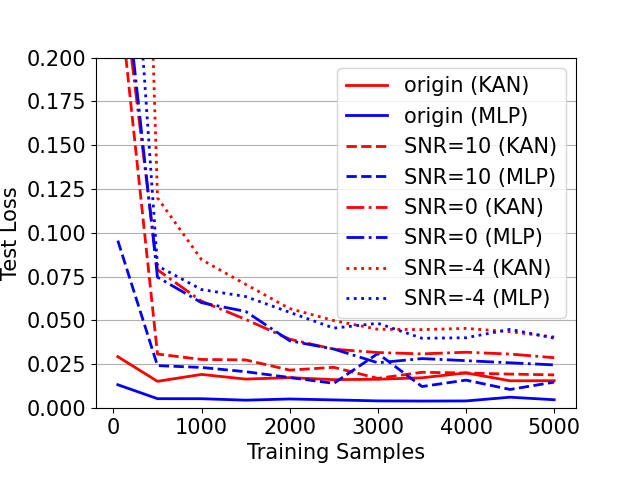}
  }

  \subfloat[\centering
  Original data, noisy data and predictions, 
  $f_3$ with noise, 
  training samples = 50 
  \label{f3_50_noise}]
  {
    \includegraphics[width=0.32\linewidth]{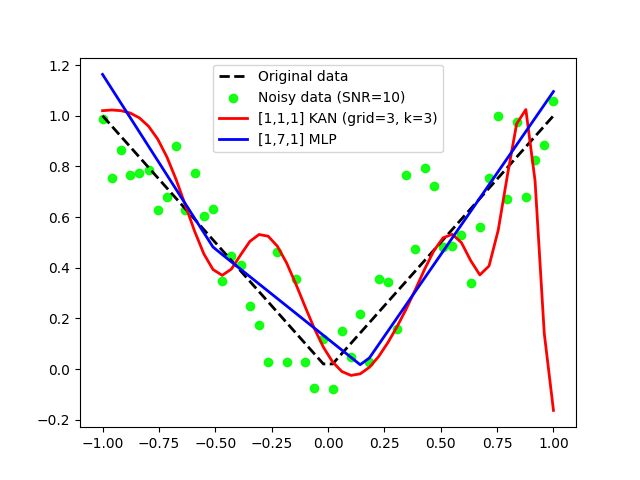}
  }
  \subfloat[\centering
  Original data, noisy data 
  and predictions of MLP, 
  $f_3$ with noise, 
  training samples = 5000 
  (500 plotted for clarity)
  \label{f3_5000_mlp_noise}]
  {
    \includegraphics[width=0.32\linewidth]{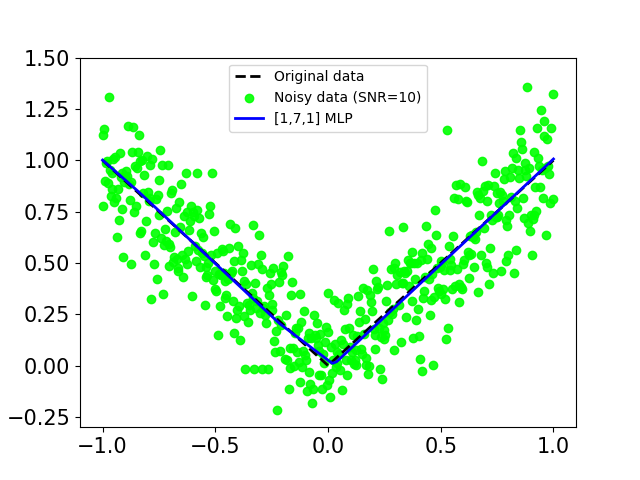}
  }
  \subfloat[\centering
  Original data, noisy data 
  and predictions of KAN, 
  $f_3$ with noise, 
  training samples = 5000 
  (500 plotted for clarity)
  \label{f3_5000_kan_noise}]
  {
    \includegraphics[width=0.32\linewidth]{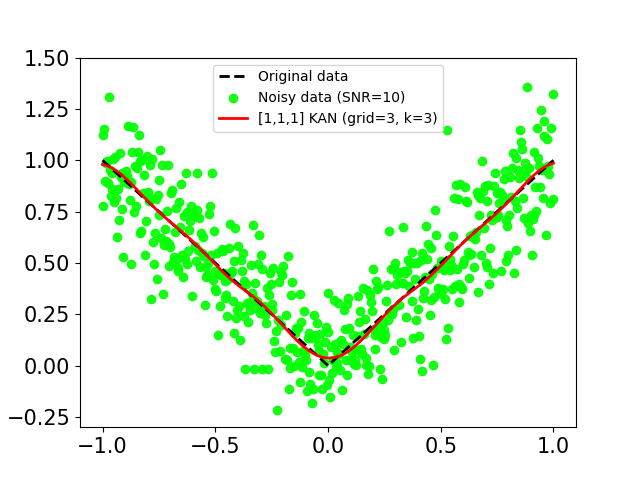}
  }

  \subfloat[\centering
  Original data, noisy data and predictions, 
  $f_4$ with noise, 
  training samples = 50 
  \label{f4_50_noise}]
  {
    \includegraphics[width=0.32\linewidth]{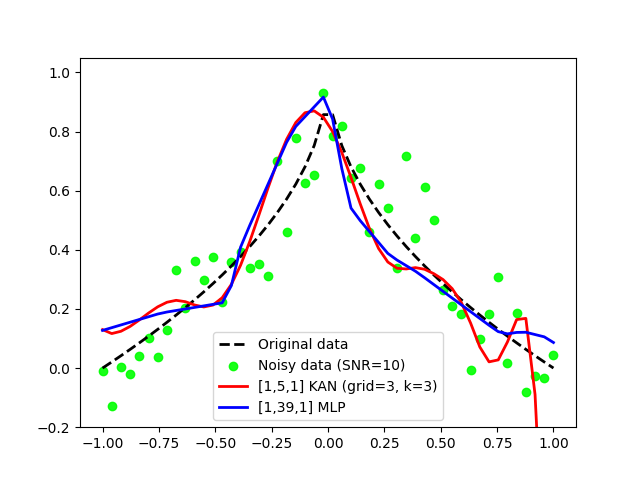}
  }
  \subfloat[\centering
  Original data, noisy data 
  and predictions of MLP, 
  $f_4$ with noise, 
  training samples = 5000 
  (500 plotted for clarity)
  \label{f4_5000_mlp_noise}]
  {
    \includegraphics[width=0.32\linewidth]{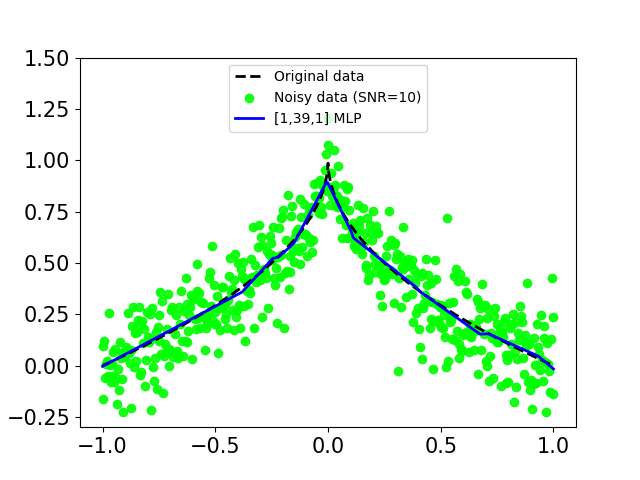}
  }
  \subfloat[\centering
  Original data, noisy data 
  and predictions of KAN, 
  $f_4$ with noise, 
  training samples = 5000 
  (500 plotted for clarity)
  \label{f4_5000_kan_noise}]
  {
    \includegraphics[width=0.32\linewidth]{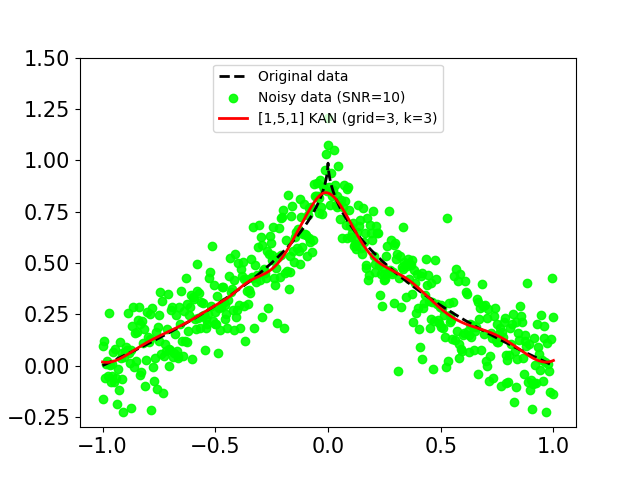}
  }

  \caption{Recover $f_3$ and $f_4$ with noise independently using KAN and MLP.}
  \label{type12_noise}
\end{figure*}

\begin{figure*}[htbp]
  \centering 
  \subfloat[\centering
  Various training samples,  
  {[1,5,1]} KAN (grid=3, k=3) and {[1,39,1]} MLP,
  $f_5$ with noise
  \label{f5_noise}]
  {
    \includegraphics[width=0.45\linewidth]{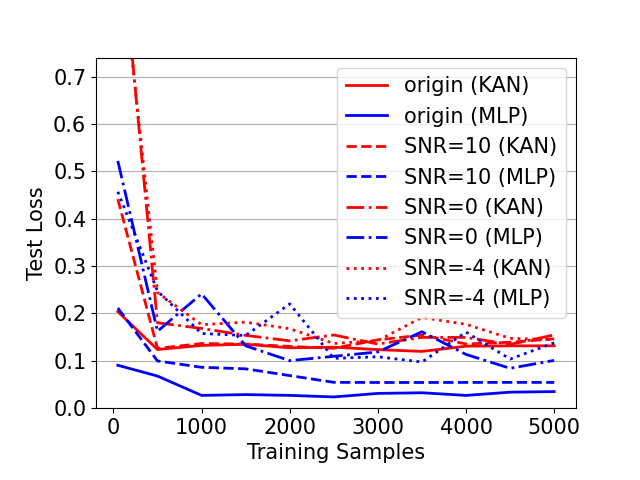}
  }
  \subfloat[\centering
  Various training samples,  
  {[1,5,1]} KAN (grid=3, k=3) and {[1,39,1]} MLP,
  $f_6$ with noise
  \label{f6_noise}]
  {
    \includegraphics[width=0.45\linewidth]{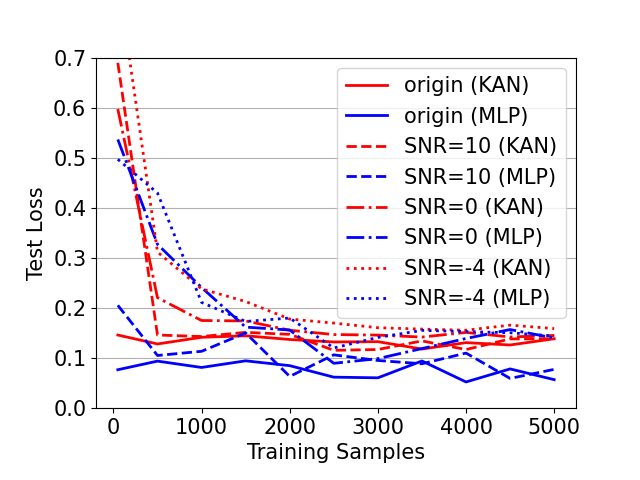}
  }

  \subfloat[\centering
  Original data, noisy data and predictions, 
  $f_5$ with noise, 
  training samples = 50
  \label{f5_50_noise}]
  {
    \includegraphics[width=0.45\linewidth]{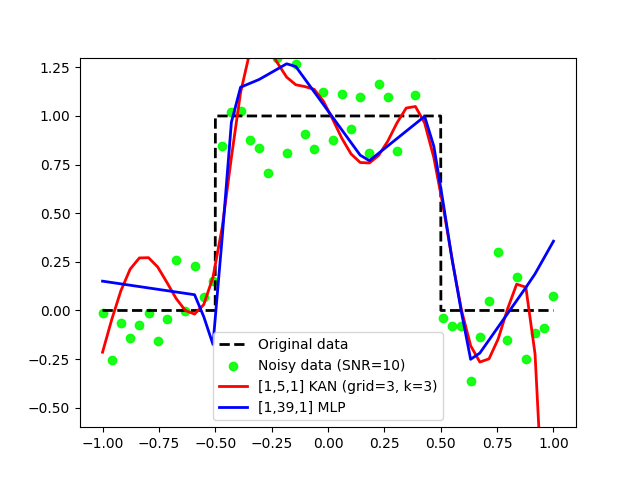}
  }
  \subfloat[\centering
  Original data, noisy data and predictions, 
  $f_5$ with noise, 
  training samples = 5000 
  (500 plotted for clarity)
  \label{f5_5000_noise}]
  {
    \includegraphics[width=0.45\linewidth]{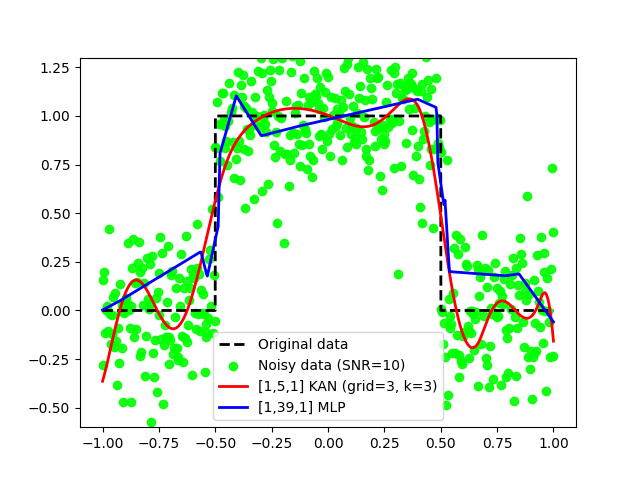}
  }

  \subfloat[\centering
  Original data, noisy data and predictions, 
  $f_6$ with noise, 
  training samples = 50
  \label{f6_50_noise}]
  {
    \includegraphics[width=0.45\linewidth]{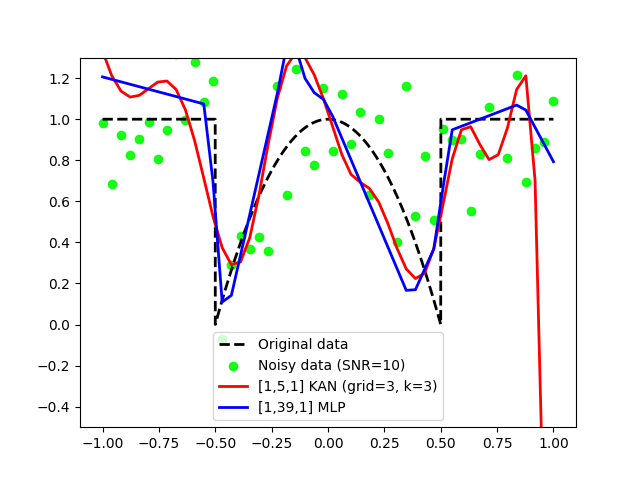}
  }
  \subfloat[\centering
  Original data, noisy data and predictions, 
  $f_6$ with noise, 
  training samples = 5000 
  (500 plotted for clarity)
  \label{f6_5000_noise}]
  {
    \includegraphics[width=0.45\linewidth]{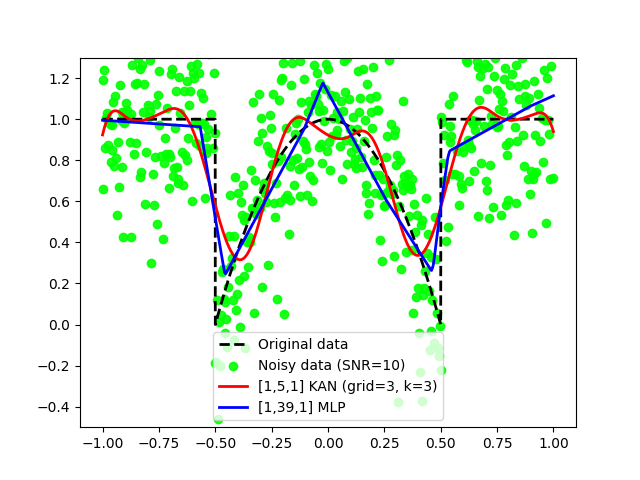}
  }
  
  \caption{Recover $f_5$ and $f_6$ with noise independently using KAN and MLP.}
  \label{type12_noise}
\end{figure*}

\end{document}